\documentclass[lettersize,journal]{IEEEtran}
\usepackage{amsmath,amsfonts}
\usepackage{algorithmic}
\usepackage{algorithm}
\usepackage{array}
\usepackage[caption=false,font=scriptsize,labelfont=sf,textfont=sf]{subfig}
\usepackage{textcomp}
\usepackage{stfloats}
\usepackage{url}
\usepackage{verbatim}
\usepackage{graphicx}
\usepackage{cite}
\usepackage{mathrsfs}
\usepackage{amsfonts}
\usepackage{amssymb}
\usepackage{multirow}
\usepackage{makecell}
\usepackage{bm}
\usepackage{xcolor}
\usepackage{booktabs}
\usepackage{subcaption}
\usepackage{subfloat}
\usepackage[normalem]{ulem}
\useunder{\uline}{\ul}{}

\begin{document}

\title{Is Precise Recovery Necessary? A Task-Oriented Imputation Approach for Time Series Forecasting on Variable Subset}

\author{Qi~Hao,~
        Runchang~Liang,~
        Yue~Gao,~
        Hao~Dong,~
        Wei~Fan,~
        Lu~Jiang,~
        Pengyang~Wang*
\thanks{Qi Hao and Pengyang Wang are with the State Key Laboratory of Internet of Things for Smart City and Department of Computer and Information Science, University of Macau, Macao SAR, China, E-mail: yc37440@um.edu.mo, pywang@um.edu.mo. 

Runchang Liang and Yue Gao are with the Northeast Normal University, Jinlin, China, Email: liangrc666@nenu.edu.cn, gaoy560@nenu.edu.cn. 

Hao Dong is with the Computer Network Information Center, Chinese Academy of Sciences, Beijing, China, Email: donghcn@gmail.com. 

Wei Fan is with the University of Oxford, London, the United Kingdom, Email: frankfanwei@outlook.com. 

Lu Jiang is with the Dalian Maritime University, Liaoning, China, Email: jiangl761@dlmu.edu.cn.}
\thanks{*Corresponding author: Pengyang Wang (email: pywang@um.edu.mo)}
}

\markboth{Journal of \LaTeX\ Class Files,~Vol.~14, No.~8, August~2021}%
{Shell \MakeLowercase{\textit{et al.}}: A Sample Article Using IEEEtran.cls for IEEE Journals}

\IEEEpubid{0000--0000/00\$00.00~\copyright~2021 IEEE}

\maketitle

\begin{abstract}
Variable Subset Forecasting (VSF) refers to a unique scenario in multivariate time series forecasting, where available variables in the inference phase are only a subset of the variables in the training phase. 
VSF presents significant challenges as the entire time series may be missing, and neither inter- nor intra-variable correlations persist. 
Such conditions impede the effectiveness of traditional imputation methods, primarily focusing on filling in individual missing data points.
Inspired by the principle of feature engineering that not all variables contribute positively to forecasting, we propose \underline{\textbf{T}}ask-\underline{\textbf{O}}riented \underline{\textbf{I}}mputation for \underline{\textbf{VSF}} (TOI-VSF), a novel framework shifts the focus from accurate data recovery to directly support the downstream forecasting task. 
TOI-VSF incorporates a self-supervised imputation module, agnostic to the forecasting model, designed to fill in missing variables while preserving the vital characteristics and temporal patterns of time series data. 
Additionally, we implement a joint learning strategy for imputation and forecasting, ensuring that the imputation process is directly aligned with and beneficial to the forecasting objective. 
Extensive experiments across four datasets demonstrate the superiority of TOI-VSF, outperforming baseline methods by $15\%$ on average.
\end{abstract}


\section{Introduction}\label{sec:intro}
\begin{figure}[!ht]
    \centering
    \includegraphics[width=0.485\textwidth]{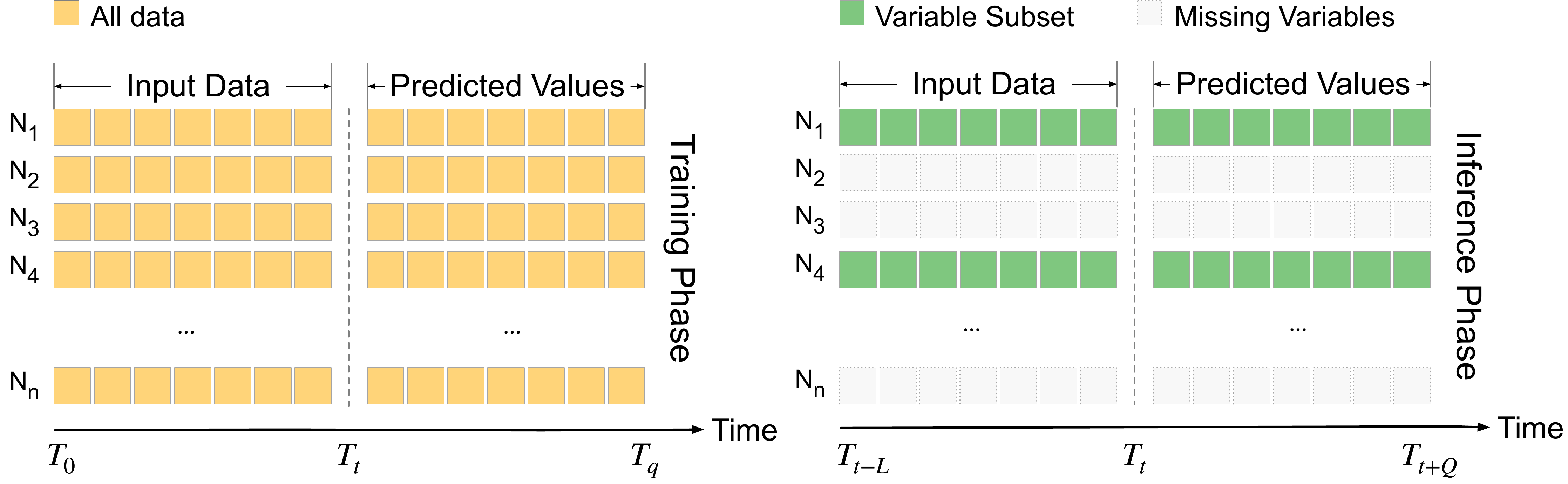}
    \caption{Variable Subset Forecast Problem: The left and the right figures show the training and inference phase respectively. The orange part represents all data in the training process. The grey part represents the missing variables. And the green part represents available subset variables.}
    \label{fig:intro1}
\end{figure}

\IEEEPARstart Variable Subset Forecasting (VSF)~\cite{10.1145/3534678.3539394} is a unique and challenging scenario in time series forecasting~\cite{fan2023dish}, where some of the variables presented during the training phase are entirely missing when making inferences, as shown in Figure~\ref{fig:intro1}. 
Unlike typical cases of missing data~\cite{wang2024deeplearningmultivariatetime}, where only a few ad hoc data points are absent, VSF involves the complete absence of certain variables during the inference phase, forcing the model to rely on a limited subset of the data it was originally trained on. 
The occurrence of VSF is often observed in contexts such as the Internet of Things (IoT)~\cite{rose2015internet}, where sensor networks are prevalent. 
For example, in a network of traffic sensors across multiple intersections~\cite{lim2021time}, certain sensors might become non-operational due to construction activities or damage, resulting in a complete loss of data from those specific locations. 
This absence of data from key nodes in the network can significantly impact the forecasting accuracy for the entire system.

\IEEEpubidadjcol

To solve the problem of VSF, an intuitive solution is data imputation, wherein missing observations are reconstructed using dependencies and correlations within and between variables~\cite{adhikari2022comprehensive}. 
Traditional imputation techniques, such as neighbor-based methods~\cite{song2015turn,sun2020swapping}, statistical methods~\cite{domeniconi2004nearest,zhang2019learning}, generative methods~\cite{tashiro2021csdi,du2023saits}, etc, rely on these intra- and inter-variable relationships. 
However, these methods prove ineffective in VSF scenarios since all the observations for certain variables are missing, rendering both intra-variable temporal dependencies and inter-variable relationships inaccessible. 
To overcome the limitation, Forecast Distance Weighting (FDW)~\cite{10.1145/3534678.3539394} has been proposed, which seeks to compensate for missing variables in the inference data by leveraging similar variables from the training dataset, thereby enhancing the completeness of the inference and leading to modest improvements in accuracy.

\begin{figure*}[!ht]
    \centering
    \includegraphics[width=\textwidth]{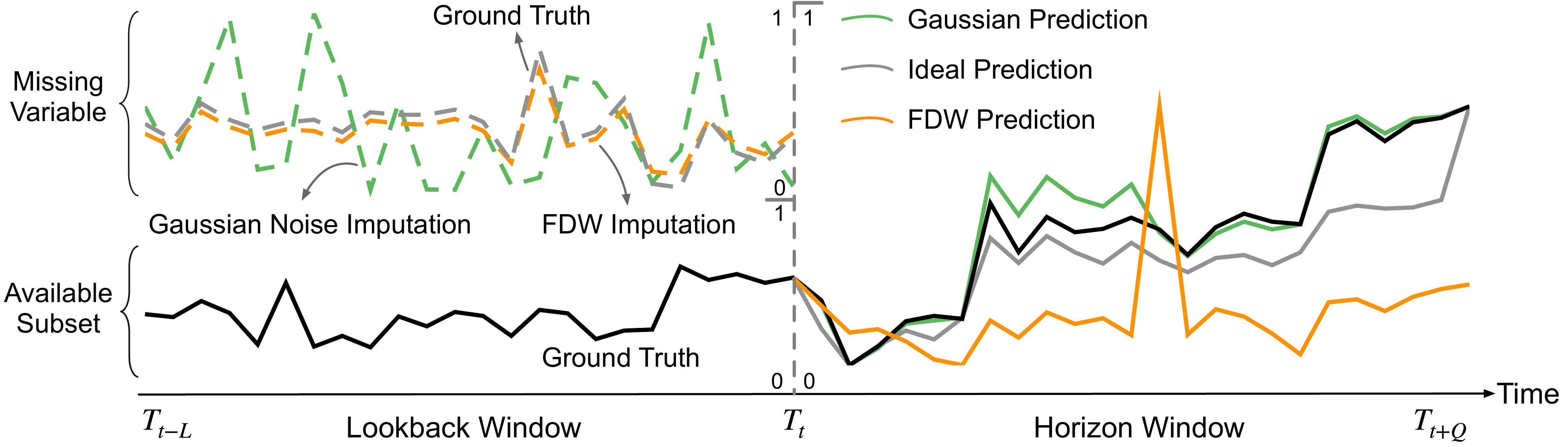}
    \caption{An example of the inference phase on the ECG5000 dataset~\cite{10.1145/3534678.3539394} utilizing the forecasting backbone MTGNN~\cite{wu2020connecting}.
    During the inference phase, only a small variable subset is available, while others are missing. 
    The objective is to obtain a precise prediction of the available variable subset.
    To simplify, we choose two variables as the example: one variable as the available variable subset (the black solid line) and one missing variable (the grey dotted line), which is negatively correlated to the available variable subset.
    To handle the missing variable, we apply different imputation strategies, including Gaussian noise-filling (the green dotted line) and traditional imputation method FDW~\cite{10.1145/3534678.3539394} (the orange dotted line).
    The forecasting performance in an ideal scenario, where no data is missing, is illustrated by the grey solid line. 
    The Gaussian imputation shows a better forecasting performance than the ideal scenario, suggesting the presence of noise or redundant information in the original data. 
    This implies that precise recovery of the missing data may not be essential for the VSF task and may even weaken the forecasting performance.}
    \label{fig:intro2}
\end{figure*}

However, a critical challenge with existing imputation strategies in VSF is their foundational goal: to replicate missing data as accurately as possible. 
From a feature engineering perspective, this approach may not always be optimal for forecasting. 
Not all variables necessarily contribute positively to the forecasting process; some may introduce bias, redundancy or negative correlations. 
In VSF, where the focus is forecasting performance on the reduced set of variables, precise imputation can inadvertently incorporate biases, redundancies, or negatively correlated information from the original data, which ultimately hampers forecasting performances. 
Figure~\ref{fig:intro2} presents an example of a well-known forecasting model MTGNN~\cite{wu2020connecting}'s performances on the variable subset (ECG5000 dataset~\cite{10.1145/3534678.3539394}), where the missing variable (grey dot line) has a negative correlation to the remaining variable (black line). 
The result indicates that even in the ideal case when the missing variable is available, the forecasting performance (purple line) is inferior to simply filling in the missing variable with the Gaussian noise (in green, dot line for filling, solid line for prediction). 
The phenomenon reveals the limitation in accurately replicating missing variables for VSF. 
Not to mention imputation cannot be perfect; the imputation error would further degrade the model performance on VSF (orange lines, FDW~\cite{10.1145/3534678.3539394}, the state-of-the-art method in VSF).

Hence, we argue that \textbf{for VSF, the objective should not be precise imputation per se, but rather an imputation strategy that is tailored for forecasting}. 
In other words, imputation for VSF is expected to align the relevance and contribution of imputed data to the forecasting process, rather than striving for exact replication of the original dataset.

To this end, we develop a novel \underline{\textbf{T}}ask-\underline{\textbf{O}}riented \underline{\textbf{I}}mputation framework for \underline{\textbf{V}}\underline{\textbf{S}}\underline{\textbf{F}} (\textbf{TOI-VSF}). 
Specifically, TOI-VSF employs a joint-learning approach that harmonizes the processes of imputation and forecasting. It operates on two fronts: 
On the one hand, the imputation model is optimized following the self-supervised learning paradigm by reconstructing randomly masked variables to maintain the essential characteristics and temporal patterns of the time series data, thus providing valuable input for the subsequent forecasting model; 
on the other hand, the loss of the forecasting model serves as the guidance for the imputation model to be regularized by the forecasting requirement. 
The proposed TOI-VSF framework is designed to be model-agnostic, making it compatible with a wide range of machine learning-based forecasting models.

In summary, our contributions are the following:
\begin{itemize}
\item\noindent We redefine imputation in VSF as a task-oriented rather than a recovery-focused challenge, offering a new perspective inspired by feature engineering principles.
\item\noindent  We propose a new variable imputation method by reconstructing masked variables via self-supervised learning.
\item\noindent We introduce a joint-learning strategy for simultaneous training of both the imputation and forecasting models, which facilitates the comprehensive assimilation of information from the data and its predictions. 
\item\noindent Empirical evaluations on four distinct datasets demonstrate the efficacy of our proposed framework in adeptly addressing the issue of missing subset variables in VSF, significantly improving the forecasting model's performance in these contexts.

\end{itemize}

\section{Problem Formulation}
Let $ \{ {x}_t^{1}, {x}_t^{2}, \cdots, {x}_t^{i}, \cdots, {x}_t^{N} \}_{t=1}^{T}$ stand for $T$-step $N$-variate time series, where ${x}_t^{i} \in \mathbb{R}^{D}$ denotes the observation of the $i$-th variable associated with $D$-dimensional features at the time step $t$. 
Considering $L$-length lookback window $\{ \mathbf{x}^{i}_{t-L:t} \}_{i=1}^{N}$, where $\mathbf{x}^i_{t-L:t} = [x^i_{t-L+1}, \cdots, x^i_{t} ]$; and subsequent $Q$-length horizon window $\{ \mathbf{x}^{i}_{t+1:t+Q}\}_{i=1}^{N}$, where $\mathbf{x}^i_{t+1:t+Q} =  [x^i_{t+1}, \cdots, x^i_{t+Q}]$, 
multivariate time series forecasting can be defined as:
\begin{align}
    \mathbf{x}^{1}_{t:t+Q},\cdots, \mathbf{x}^{N}_{t:t+Q} = \mathscr{F}_\Theta \left(\mathbf{x}^{1}_{t-L:t},\cdots, \mathbf{x}^{N}_{t-L:t} \right),
    \label{eq:base_ar}
\end{align}
where $\mathscr{F}_{\Theta}:  \mathbb{R}^{L \times N \times D} \rightarrow  \mathbb{R}^{Q \times N \times D}$ denotes the forecasting model parameterized by $\Theta$.

\noindent \textbf{Variable Subset Forecasting (VSF)}.
During the inference phase, only an arbitrary variable subset of the original data is available. 
Let $\Psi_N$ denote the original variable space, $\Psi_S$ denote the available subset with $S$ variables in the inference phase, where $\Psi_S \subset \Psi_N$ and $S \ll N$. 
Then, the available observations for the forecasting model $\mathscr{F}_\Theta$ is in $\mathbb{R}^{L \times S \times D}$. 
The objective of VSF is to enhance the capability of the forecasting model $\mathscr{F}_\Theta$, which was trained on the complete variable set $\Psi_N$, to generalize effectively on the subset $\Psi_S$.

\section{Methodology}
\begin{figure*}[!t]
    \centering
    \includegraphics[width=1\textwidth]{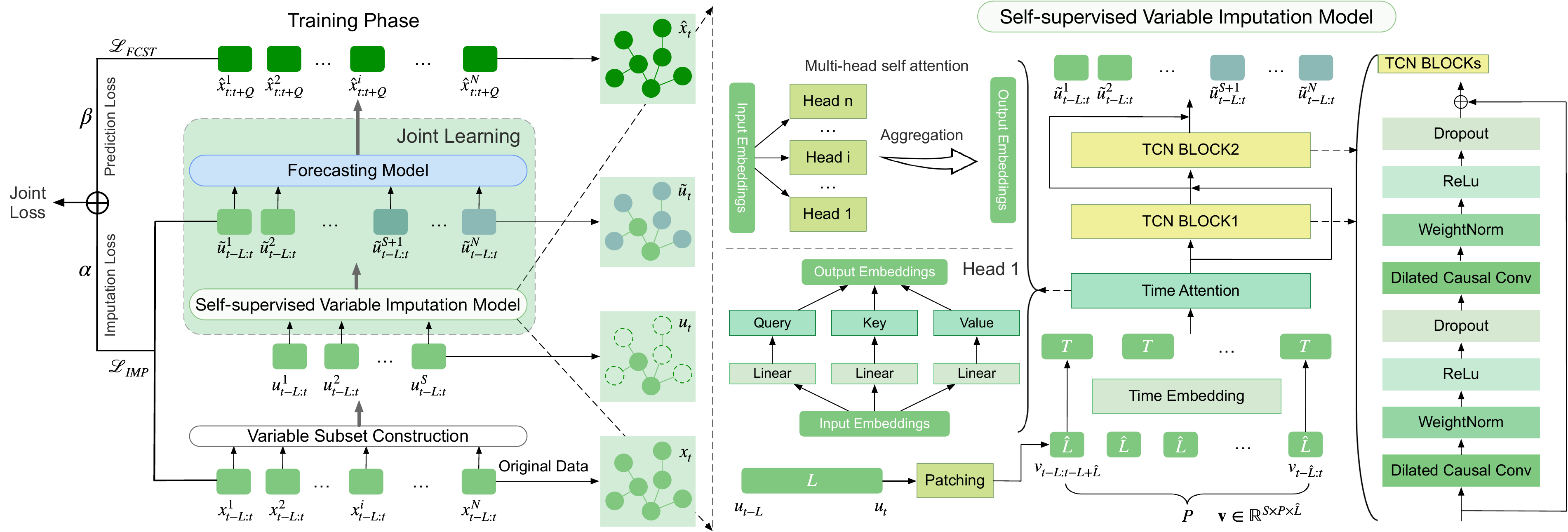}
    \caption{Framework Overview. \textbf{Left:} The training phase. 
    For multivariate time series $\{ \mathbf{x}^{i}_{t-L:t} \}_{i=1}^{N}$, certain variables $N-S$ are randomly masked to obtain a variable subset $\{ \mathbf{u}^{i}_{t-L:t} \}_{i=1}^{S}$.
    The subset $\Psi_S$ is then fed to the self-supervised learning model to obtain a reconstructed time series $\{ \mathbf{\tilde{u}}^{i}_{t-L:t} \}_{i=1}^{N}$. Subsequently, the reconstructed time series is used for the forecasting task, ultimately yielding the prediction results $\{ \mathbf{\hat{x}}^{i}_{t:t+Q} \}_{i=1}^{N}$. 
    \textbf{Right:} The self-supervised variable imputation model.
    The reconstructed time series $\{ \mathbf{\tilde{u}}^{i}_{t-L:t} \}_{i=1}^{N}$ is obtained by feeding the input variable subset $\{ \mathbf{u}^{i}_{t-L:t} \}_{i=1}^{S}$ into the model.
    }
    \label{fig: framework}
\end{figure*}

In this section, we present our proposed framework, TOI-VSF. 
The proposed TOI-VSF breaks the boundary of imputation and forecasting, and conducts better imputation for VSF with forecasting as guidance. 
Specifically, as shown in Figure~\ref{fig: framework}, TOI-VSF includes two phases: 
(1) Phase I, self-supervised imputation for completing missing variables; 
(2) Phase II, imputation-forecasting joint learning to equip imputed variables with better characteristics benefiting forecasting. 
Next, we introduce TOI-VSF in detail.

\subsection{Phase I: Self-supervised Variable Imputation} 
Given that variable subsets can vary arbitrarily during the inference phase, it is crucial for the variable imputation model to effectively generalize across all potential combinations of variables. 
To address this, we develop a self-supervised module for completing missing variables. 
The module includes five key components, including:
(1) subset construction, 
(2) patching,
(3) time embedding, 
(4) time attention, and 
(5) variable generation.

\subsubsection{Subset Construction}
To support the self-supervised imputation, we first construct various variable subsets as the training data.
Specifically, given a time period, a variable subset $\Psi_S$ is sampled by randomly masking a portion of the variables. We set $S$ to be some percentage $k$ of $N$.
It is important to note that the subset sampling is random, and multiple rounds of sampling are performed to ensure its broad coverage for the possible variable combinations. 
To distinguish the variable subsets from the complete data, for a given subset, we replace all the $x$s with $u$s to denote the data in the subset. 
Then, the data in the sampled subsets can be denoted as 
\begin{align}
\text{Lookback: }\{ \textbf{u}^{j}_{t-L:t}  | \mathbf{u}^j_{t-L:t} = [u^j_{t-L+1}, \cdots, u^j_{t} ]\}_{j=1}^{S}, \\
\text{Horizon: }\{\mathbf{u}^{j}_{t:t+Q} | \mathbf{u}^j_{t:t+Q} =  [u^j_{t+1}, \cdots, u^j_{t+Q}]\}_{j=1}^{S}, 
\end{align} 
where $u_{\ast}^{j}$ denotes the observation of the $j$-th variable the subset at a given time step, and $\mathbf{u}^j_{\ast} \in \Psi_S$.

\subsubsection{Patching}
We divide the input time series $\{ \mathbf{u}^{i}_{t-L:t} \}_{i=1}^{S}$ into $P$ patches which are non-overlapped.
So, we will generate a sequence of patches $\mathbf{v}^{i} \in \mathbb R ^{P\times \hat{L}}$, where $\hat{L}$ is the patch length.
And the number of input lengths will reduce from $L$ to $L/P$.
During the patching stage, not only do we reduce computational complexity and memory storage requirements, but we also place greater emphasis on local features, thereby enhancing the model's generalization capabilities.

\subsubsection{Time Embedding}
Then we employ a linear layer to capture the temporal patterns within the series of each variable. 
Specifically, given the sampled subset $\{ \mathbf{v}^{j}_{t-L:t}\}_{j=1}^{S}\in \mathbb{R}^{S\times P \times \hat{L}}$ as input, the linear operation is performed on the time dimension. 
Then, the projected embedding $\{ \mathbf{e}^{j}_{t-L:t}\}_{j=1}^{S}\in \mathbb{R}^{S\times P \times T}$ can be denoted as 
\begin{equation}
    \{ \mathbf{e}^{j}\}_{j=1}^{S} = \text{Linear}(\{\mathbf{v}^{j}\}_{j=1}^{S}),
    \label{equ:embedding}
\end{equation}
where $\text{Linear}: \mathbb R^{S\times P\times \hat{L}} \rightarrow  \mathbb{R}^{S \times P \times T}$ denotes the time embedding function.

\subsubsection{Time Attention}
For accurate imputation, it is not enough to capture the temporal features.
We employed an attention module to enhance our understanding of the interrelationships between variables.
Specifically, given the input $\{\textbf{e}^{j}\}_{j=1}^{S} \in \mathbb R^{S*P*T}$ 
, we directly apply multi-head self-attention (MSA) to each variable $\mathbf{e}^{j}$, $j\in[1, S]$:
\begin{align}
    \mathbf{z}^{j} &= \text{LayerNorm}(\mathbf{e}^{j} + \text{MSA}(\mathbf{e}^{j},\mathbf{e}^{j},\mathbf{e}^{j}))\\
    {\mathbf{\hat{z}}^{j}} &=\text{LayerNorm}(\mathbf{z}^{j}+\text{MLP}(\mathbf{z}^{j})),
\end{align}
where $\text{LayerNorm}$ denotes a commonly adopted layer normalization~\cite{vaswani2017attention}; 
$\text{MLP}$~\cite{zhou10beyond} stands for a multi-layer feedforward network; 
$\text{MSA}(Q,K,V)$ represents the multi-head self-attention layer~\cite{vaswani2017attention} where $Q$, $K$ and $V$ correspond to queries, keys and values, respectively. 
Note that all variables share the same MSA layer.

\subsubsection{Variable Generation}
We develop a Temporal Convolutional Network (TCN)~\cite{bai2018empirical} architecture for generating variables (time series).
Specifically, the architecture includes two blocks, where each block is a causal convolution, with Rectified Linear Unit (ReLU) activation in between.
The first convolution layer is a mapping of the vector, and the second is the representation of the mapped vector.
The block can be denoted as 
\begin{align}
\text{Conv} = \sum_{k=0}^{K-1} \mathbf{W}_k \cdot \mathbf{\hat{z}}_{t-k}+\mathbf{b}_1,
\end{align}
where $K$ is the kernel size; 
$d$ is the dilation factor; 
$\mathbf{W}$ is the convolution kernel matrix; 
$\mathbf{b}_{1}$ is the bias. 
And $t-d\cdot k$ accounts for the direction of the past.
\begin{equation}
    \text{BLOCK}(\cdot) = \text{ReLU}(\text{Conv}(\text{ReLU}(\text{Conv}(\cdot)))).
\end{equation}

Then, the two blocks are linked with a residual connection, followed by a linear prediction layer. 
The workflow can be denoted as
\begin{align}
    \mathbf{A} &= \text{BLOCK}_1(\{ \mathbf{\hat{z}}^{j}_{t-L:t}\}_{j=1}^{S}), \\
    \mathbf{B} &= \text{BLOCK}_2(\{ \{ \mathbf{\hat{z}}^{j}_{t-L:t}\}_{j=1}^{S} + \mathbf{A}),    
\end{align}
\begin{equation}
    \{ \mathbf{\tilde{u}}^{i}_{t-L:t} \}_{i=1}^{N} = \text{GeLU}(\mathbf{w}_1\mathbf{B}+\mathbf{b}_2), 
\end{equation}
where $\mathbf{w}_{1}$ and $\mathbf{b}_{2}$ represent the weights and bias terms. $\{ \mathbf{\tilde{u}}^{i}_{t-L:t} \}_{i=1}^{N}$ denotes the generated series for $N$ variables.

The objective of self-supervised training is to minimize the differences between the groud-truth $\{ \mathbf{x}^{i}_{t-L:t} \}_{i=1}^{N}$ and the generated results $\{ \mathbf{\tilde{u}}^{i}_{t-L:t} \}_{i=1}^{N}$. 
Therefore, the loss of the proposed self-supervised variable imputation can be represented in the format of Mean Absolute Error (MAE) as
\begin{equation}
    \mathcal{L}_{\text{IMP}} = \frac{1}{N} \sum \limits_{i=1}^{N} |\mathbf{x}^{i}_{t-L:t} - \mathbf{\tilde{u}}^{i}_{t-L:t}|.
\end{equation}

\begin{figure}[!t]
    \centering
    \includegraphics[width=0.47\textwidth]{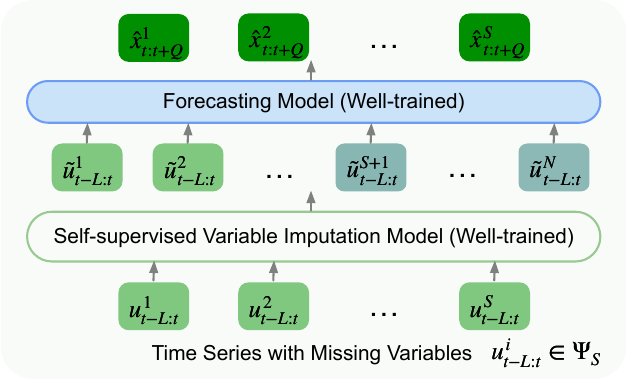}
    \caption{Inference Phase. 
    Different from the training phase, the test data only contains a variable subset $\{ \mathbf{u}^{i}_{t-L:t} \}_{i=1}^{S}$. 
    By the trained model, we can generate the new time series $\{ \mathbf{\tilde{u}}^{i}_{t-L:t} \}_{i=1}^{S}$ and get more accurate predictions $\{ \mathbf{\hat{x}}^{i}_{t:t+Q} \}_{i=1}^{S}$. 
    Notably, the performance of VSF is only evaluated on the variable subset $\Psi_S$.}
    \label{fig: inference}
\end{figure}

\subsection{Phase II: Imputation-Forecasting Joint Learning}
As discussed in the Introduction, precise recovery of data is unnecessary for the forecasting task, even bringing bias and negative impacts on subset forecasting. 
Therefore, it is intuitive to take forecasting as guidance to regularize the variable imputation model toward better forecasting performance. 

Specifically, let $\{\mathbf{\hat{x}}^{i}_{t:t+Q}\}_{i=1}^{N}$ denote the predicted results derived by the forecasting model $\mathscr{F}_\Theta$ by taking the generated variables $\{\mathbf{\tilde{u}}^{i}_{t-L:t}\}_{i=1}^{N}$ as input, that is
\begin{equation}
    \{\mathbf{\hat{x}}^{i}_{t:t+Q}\}_{i=1}^{N} = \mathscr{F}_\Theta(\{\mathbf{\tilde{u}}^{i}_{t-L:t}\}_{i=1}^{N}).
\end{equation}
Then, the forecasting loss can be represented as 
\begin{equation}
    \mathcal{L}_{\text{FCST}} = \frac{1}{N} \sum\limits_{i=1}^{N} |\{\mathbf{x}^{i}_{t:t+Q}\} - \{\mathbf{\tilde{x}}^{i}_{t:t+Q}\}|.
\end{equation} 

Different from the traditional data imputation method, we propose a joint learning strategy to train the variable imputation model and the forecasting model concurrently from the task-oriented perspective. 
As the parameters of the generation module and forecasting module mutually influence each other, simultaneous training enables a more effective optimization of both modules, leading to more accurate predictive results, as elaborated in Section~\ref{joint-learning}.

Therefore, our objective function contains two parts, the self-supervised variable imputation loss $\mathcal{L}_{\text{IMP}}$ and the forecasting loss $\mathcal{L}_{\text{FCST}}$.
The self-supervised variable imputation loss $\mathcal{L}_{\text{IMP}}$ is to ensure that the generated features maintain temporal patterns rather than generating meaningless numerical values, 
while the forecasting loss $\mathcal{L}_{\text{FCST}}$ incentivizes the imputation model to generate missing variables in a manner that optimizes forecasting performance.
By combining the pre-training loss and prediction loss, we ultimately obtain our loss function:
\begin{equation}
    \mathcal{L} = \alpha \cdot \mathcal{L}_{\text{IMP}} + \beta \cdot  \mathcal{L}_{\text{FCST}},
    \label{total-loss}
\end{equation}
where $\alpha$ and $\beta$ are hyperparameters to control the contribution of imputation and forecasting models, respectively, and $\alpha + \beta = 1$. 
We optimize the proposed framework by minimizing the joint learning loss $\mathcal{L}$ via gradient descent. 

\subsection{Inference on Variable Subset}

In the inference phase, we adapt the well-trained TOI-VSF to the new variable subset for forecasting. 
Here, the new variable subset will be taken as the input to 
the patching phase, and facilitate the execution of variable imputation as depicted in Figure~\ref{fig: inference}.
Then, the generated complete variables are employed as the input for the forecasting model to conduct inference.
We collect predicted results on the subset $\Psi_{S}$ as the final output for the VSF task.
\section{Experiment}
In this section, extensive experiments are conducted to evaluate the effectiveness of the proposed model.
Our experiments aim to answer the following research questions (\textbf{RQ}): 
\begin{itemize}
\item\noindent \textbf{RQ1:} 
Does our proposed model actually work on the problem of variable subset forecasting? 
\item\noindent \textbf{RQ2:} 
Can our proposed task-oriented imputation method break the upper boundary of VSF on the complete variable set (an ideal scenario with no missing data)?
\item\noindent \textbf{RQ3:} Does joint learning truly benefit the forecasting performance on variable subset?
\item\noindent \textbf{RQ4:} How do the weights of imputation and forecasting loss impact the final forecasting performance?
\item\noindent \textbf{RQ5:} Can the proposed model outperform the traditional data imputation methods in variable subset forecasting?
\item\noindent \textbf{RQ6:}
How does the size of the available variable subset with different $k$ affect the final forecasting performance?
\end{itemize}

\subsection{Experimental Setup}\label{setup}
\noindent\textbf{Datasets.}\quad 
We conduct our experiments on four real-world datasets:
(i) \textbf{\textit{METR-LA\footnote{https://www.kaggle.com/datasets/annnnguyen/metr-la-dataset}}} dataset 
covers the average traffic speed of the highways in Los Angeles County, collected from 207 loop detectors from Mar 2012 to Jun 2012.
(ii) \textbf{\textit{TRAFFIC\footnote{https://pems.dot.ca.gov/}}} dataset records the road occupancy rates measured by 862 sensors in the San Francisco Bay area during 2015 and 2016. 
Since a significant portion of the values in the dataset is on the order of $1e^{-3}$ by default, we enhance the scale by multiplying the variable values by $1e^3$.
(iii) \textbf{\textit{SOLAR\footnote{https://www.nrel.gov/grid/solar-power-data.html}}} dataset records the solar power output collected from 137 plants in Alabama State in 2007.
(iv) \textbf{\textit{ECG5000\footnote{http://www.timeseriesclassification.com/description.php?Dataset=ECG5000}}} dataset, sourced from the UCR Time-Series Classification Archive, comprises 140 electrocardiograms (ECG) with a length of 5000 each.
Similar to the Traffic dataset, we increase the scale of all variables by a factor of 10.
(v) \textbf{\textit{ETTH1}} dataset~\cite{haoyietal-informer-2021}, 
spans two years from July 2016 to July 2018, capturing the load characteristics of seven oil and power transformers related to power transmission, with data granularity at an hourly level.
\noindent\textbf{Backbones.}\quad 
The self-supervised variable imputation model is a generic framework that can be integrated into various multivariate time series forecasting models. 
We conduct the evaluation in the ablation study manner, by coupling our TOI-VSF with four widely-used backbone forecasting models,
MTGNN~\cite{wu2020connecting}, ASTGCN~\cite{guo2019attention}, MSTGCN~\cite{guo2019attention} and TGCN~\cite{zhang2021tgcn}.
\textbf{\textit{MTGNN}}~\cite{wu2020connecting} utilizes a graph learning module to extract one-way relationships between variables, followed by joint learning of spatio-temporal relationships for spatio-temporal modelling. 
\textbf{\textit{ASTGCN}}~\cite{guo2019attention} incorporates adaptive weighting through the use of advanced spatiotemporal attention mechanisms, which are designed to dynamically discern and learn the complexities of spatiotemporal relationships, thereby enhancing the accuracy and efficacy of predictive models.
\textbf{\textit{MSTGCN}}~\cite{guo2019attention} utilizes a multiscale graph convolutional network to capture spatial features and employs temporal convolution along the time dimension to capture temporal dependencies.
\textbf{\textit{TGCN}}~\cite{zhang2021tgcn} introduces temporal feature learning to capture the evolution of time features and employs graph convolutional networks to capture the spatio-temporal relationships between nodes.

\noindent\textbf{Data Imputation Baselines.}\quad
To fully validate the performance of the proposed TOI-VSF, we compare it with the soda baseline in VSF, FDW~\cite{10.1145/3534678.3539394}, and also the below data imputation baselines, KNNE~\cite{domeniconi2004nearest}, IIM~\cite{zhang2019learning}, TRMF~\cite{yu2016temporal}, CSDI~\cite{tashiro2021csdi},
SAITS~\cite{du2023saits},
SS-GAN~\cite{miao2021generative} and 
MICE~\cite{van2011mice}.
A detailed introduction to these methods will be explained in Section 5.1.

\begin{table*}[!t]
\small
\setlength\tabcolsep{1.8pt}
\caption{An illustration of the overall performance of time series forecasting on subset variables. 
The results are reported as an average and standard deviations (values in parenthesis) of 10 runs. 
Each run corresponds to one trained model with 100 epochs, and each epoch involves a randomly sampled variable subset.
$\Delta_{subset}$ represents the improvement of TOI-VSF over the \textit{Partial} setting. For the $\Delta_{subset}$, the larger the value, the better the performance.}
\begin{tabular}{c|c|cc|cc|cc|cc|cc}
\toprule
\multicolumn{2}{c|}{\multirow{2.7}{*}{Forecasting Model}  }        & \multicolumn{2}{c|}{METR-LA}                 & \multicolumn{2}{c|}{TRAFFIC}                 & \multicolumn{2}{c|}{SOLAR}                   & \multicolumn{2}{c|}{ECG5000}         & \multicolumn{2}{c}{ETTH1}         \\

\cmidrule{3-12}
\multicolumn{2}{c|}{}                                       & MAE                  & RMSE                 & MAE                  & RMSE                 & MAE                  & RMSE                 & MAE                  & RMSE            & MAE                  & RMSE     \\
\midrule
\multirow{5}{*}{ASTGCN}         & \textit{Partial}                   & 5.57(0.72)           & 10.61(1.36)          & 22.44(1.58)          & 43.07(2.46)          & 6.14(1.29)           & 8.95(2.35)           & 3.60(0.60)           & 6.05(1.13)       & 2.21(1.10)          & 3.44(1.98)   \\
& \textit{Oracle}                                         & 5.04(0.39)           & 9.59(0.62)           & 19.17(0.91)          & 40.21(2.02)          & 4.54(0.47)           & 6.48(0.85)           & 3.47(0.50)           & 5.83(0.99)          & 1.98(1.09)          & 3.27(1.93) \\
& TOI-VSF                                             & \textbf{4.34(0.28)}  & \textbf{8.13(0.59)}  & \textbf{18.90(1.04)} & \textbf{39.69(2.28)} & \textbf{3.14(0.42)}  & \textbf{4.74(0.68)}  & \textbf{3.18(0.46)}  & \textbf{5.40(1.08)}  & \textbf{1.87(1.16)} & \textbf{3.23(2.07)}\\

\cmidrule{2-12}
& {$\Delta_{subset}$}     & 22.08\% & 23.37\% & 15.78\% & 7.85\%  & 48.86\% & 47.04\% & 11.67\% & 10.74\% & 15.38\% & 6.10\%\\
\midrule
\multirow{5}{*}{MSTGCN}         & \textit{Partial}                   & 4.78(0.43)           & 9.35(0.75)           & 18.96(1.21)          & 40.13(2.67)          & 4.75(0.73)           & 7.02(1.42)           & 4.43(0.87)           & 7.61(1.86)           & 2.22(1.08)          & 3.44(1.93) \\
& \textit{Oracle}                                           & 4.49(0.31)           & 8.93(0.50)           & 17.41(0.74)          & 37.84(1.88)          & 3.64(0.41)           & 5.60(0.82)           & 3.39(0.52)           & 5.82(1.06)           & 1.91(1.05)          & 3.17(1.86)\\
& TOI-VSF                                             & \textbf{3.91(0.24)}  & \textbf{7.66(0.44)}  & \textbf{17.36(0.84)} & \textbf{37.65(2.20)} & \textbf{2.56(0.25)}  & \textbf{4.04(0.54)}  & \textbf{3.18(0.50)}  & \textbf{5.41(0.96)}  & \textbf{1.86(1.08)} & \textbf{3.12(1.95)}\\

\cmidrule{2-12}
& {$\Delta_{subset}$}     & 18.20\% & 18.07\% & 8.44\%  & 6.18\%  & 46.11\% & 42.45\% & 28.22\% & 28.91\% & 12.61\% & 5.81\%\\
\midrule
\multirow{5}{*}{MTGNN}          & \textit{Partial}                   & 4.54(0.37)           & 8.9(0.68)            & 18.57(2.31)          & 38.46(3.94)          & 4.36(0.53)           & 6.04(0.81)           & 3.88(0.61)           & 6.54(1.10)           & 2.02(1.17)          & 3.34(2.01)\\
& \textit{Oracle}                                          & 3.49(0.25)           & 7.21(0.50)           & 11.45(0.57)          & 27.48(2.14)          & 2.94(0.27)           & 4.66(0.57)           & 3.43(0.54)           & 5.94(1.08)           & 1.75(0.90)          & 2.89(1.61)\\
& TOI-VSF                                             & \textbf{3.28(0.21)} & \textbf{6.51(0.38)}  & \textbf{11.40(0.61)} & \textbf{27.35(2.02)} & \textbf{2.26(0.23)} & \textbf{3.53(0.47)} & \textbf{3.21(0.49)} & \textbf{5.51(0.93)} & \textbf{1.72(0.95)} & \textbf{2.88(1.68)}\\

\cmidrule{2-12}
& {$\Delta_{subset}$}     & 27.75\% & 26.85\% & 38.61\% & 28.89\% & 48.17\% & 41.56\% & 17.27\% & 15.75\% & 14.85\% & 13.77\%\\
\midrule
\multirow{5}{*}{TGCN}           & \textit{Partial}                  & 9.92(0.75)          & 15.66(0.94)          & 43.43(1.89)          & 68.72(2.90)          & 8.76(0.87)          & 12.51(1.63)          & 6.22(1.37)          & 9.91(2.27)          & 3.75(1.65)          & 5.54(2.62)\\
& \textit{Oracle }                                         & 8.57(0.92)          & 14.78(1.27)          & 30.09(1.32)          & 53.58(2.62)          & 4.56(0.78)          & 7.32(1.64)          & 6.16(1.29)          & 9.84(2.20)          & 3.69(1.55)          & 5.44(2.44)\\
& TOI-VSF                                             & \textbf{8.46(0.94)} & \textbf{14.14(1.28)} & \textbf{28.82(1.46)} & \textbf{51.14(2.75)} & \textbf{4.43(0.78)} & \textbf{7.15(1.57)} & \textbf{6.13(1.42)} & \textbf{9.63(2.36)} & \textbf{3.60(2.03)} & \textbf{5.32(2.46)}\\

\cmidrule{2-12}
& {$\Delta_{subset}$}     & 14.72\% & 9.71\%  & 33.64\% & 25.58\% & 49.43\% & 41.15\% & 1.50\%  & 2.81\%  & 4.00\%  & 3.97\%\\
\midrule
\midrule
\multicolumn{2}{c|}{Average Improvement}                                             & \textbf{20.69\%} & \textbf{19.50\%} & \textbf{24.12\%} & \textbf{17.12\%} & \textbf{48.14\%} & \textbf{43.05\%} & \textbf{14.66\%} & \textbf{14.55\%} & \textbf{12.61\%} & \textbf{8.29\%} \\

\bottomrule
\end{tabular}
\label{tab: overall performance}
\end{table*}

\noindent\textbf{Comparison Setting.}\quad
In addressing the problem of time series forecasting on subset variables,  we assume during the training stage that the data is complete, while the inference phase differs from the training stage. 
During inference, the input data $\{ \mathbf{u}^{j}_{t-L:t} \}_{j=1}^{S}$ consists of only a subset of variables $\Psi_S$, a portion of all variables $\Psi_N$. 
To generate more accurate forecasting results, we first impute the missing variables, and then feed them as input to the trained forecasting model.
We evaluate the forecast performance of the model only on subset variables $\Psi_S$.

We implement two different settings for each of the four forecasting backbones described below. 
For two settings, the training process is identical, utilizing the entire data. 
The only difference lies in their respective inference stages.
\begin{itemize}
\item\noindent \textit{\textbf{Partial.}} We completely remove the data for the variables $N-S$ during the test stage and do not perform any data imputation for these missing variables. 
This reflects the original performance of the forecasting model on Variable Subset Forecasting (VSF).

\item\noindent \textit{\textbf{Oracle.}} Oracle is an ideal scenario of VSF problem only for comparison. 
Specifically, we assume that all $N$ variables data is available during the inference phase.  
Notably, the loss metrics are computed only for the variables in $\Psi_S$, aligning with our comparison objectives.
\end{itemize}

\noindent\textbf{Evaluation Metrics.}\quad 
We evaluate the performances using two commonly used metrics in multivariate time series forecasting, including Mean Absolute Error (MAE) and Root Mean Squared Error (RMSE).
In addition, we use two new metrics to compare the performance of our method and the two above different settings.

\begin{itemize}
\item\noindent \textit{\textbf{Partial Improvement}} $(\Delta_{subset})$.
We use $\Delta_{subset}$ to quantify the performance of the proposed method for the \textit{Partial} setting, defined as the follows: 
\begin{align}
\Delta_{subset} = {\frac{E_{partial} - E_{ours}}{E_{partial}} } \times 100\%.
\end{align}
Here, $E$ serves as the placeholder for the error metrics MAE and RMSE. 
$E_{partial}$ are errors for the \textit{Partial} setting and $E_{ours}$ represents the errors for TOI-VSF.

\item\noindent \textbf{\textit{Oracle Improvement}$(\Delta_{improve})$.}
We compare our proposed method with the \textit{Oracle} setting. 
Let $E_{oracle}$ denote the error of the oracle setting, the improvement $\Delta_{improve}$ can be represented as 
\begin{align}
\Delta_{improve} = {\frac{E_{oracle} - E_{ours}}{E_{oracle}} } \times 100\%.
\end{align}
\end{itemize}

For the improvement measurements $\Delta_{subset}$ and $\Delta_{improve}$, the larger the value, the better the performance.

\begin{table*}[!h]

\centering
\small
\setlength\tabcolsep{7.3pt}
\caption{Relative performance to \textit{Oracle} setting. $\Delta_{improve}$ indicates the proportion by which TOI-VSF surpasses the \textit{Oracle} setting. For the $\Delta_{improve}$, the larger the value, the better the performance.}
\begin{tabular}{c|c|cc|cc|cc|cc|cc}
\toprule
\multicolumn{2}{c|}{\multirow{2.7}{*}{Forecasting Model}  }& \multicolumn{2}{c|}{METR-LA} & \multicolumn{2}{c|}{TRAFFIC} & \multicolumn{2}{c|}{SOLAR} & \multicolumn{2}{c|}{ECG5000} & \multicolumn{2}{c}{ETTH1}\\
\cmidrule{3-12}
\multicolumn{2}{c|}{}& MAE  & RMSE    & MAE  & RMSE  & MAE  & RMSE & MAE  & RMSE   & MAE  & RMSE\\
\midrule
ASTGCN  & $\Delta_{improve}$ & 13.89\%         & 15.22\%          & 1.41\%          & 1.29\%          & 30.84\%          & 26.85\%          & 8.36\%          & 7.38\%          & 5.56\%          & 1.22\%\\
\midrule
MSTGCN  & $\Delta_{improve}$ & 12.92\%         & 14.22\%          & 0.29\%          & 0.50\%          & 29.67\%          & 27.86\%          & 6.19\%          & 7.04\%          & 2.62\%         & 1.58\%\\
\midrule
MTGNN  & $\Delta_{improve}$ & 6.02\%          & 9.71\%           & 0.44\%          & 0.47\%          & 23.13\%          & 24.25\%          & 6.41\%          & 7.24\%          & 1.71\%          & 0.35\%\\
\midrule
TGCN  & $\Delta_{improve}$ & 1.28\%          & 4.33\%           & 4.22\%          & 4.55\%          & 2.85\%           & 2.32\%           & 0.52\%          & 2.17\%          & 2.44\%          & 2.21\%\\
\midrule
\midrule
\multicolumn{2}{c|}{Average Improvement}& \textbf{8.53\%} & \textbf{10.87\%} & \textbf{1.59\%} & \textbf{1.71\%} & \textbf{21.62\%} & \textbf{20.32\%} & \textbf{5.37\%} & \textbf{5.96\%} & \textbf{3.08\%} & \textbf{1.34\%}\\
\bottomrule

\end{tabular}
\label{tab:oracle-ours}
\end{table*}

\noindent\textbf{Implementation Details.}\quad 
During the inference phase, only a subset of variables $\Psi_S$ is available. 
We set $S$ to be some percent $k$ of all variables $N$, $k=15\%$ in this paper. 
For the training phase, we randomly sample the subset $\Psi_S$ 100 times, ensuring its broad coverage within the complete data $\Psi_N$. 
We conduct 10 times training leading to 10 trained models, where each model is trained with 100 epochs, and each epoch takes one randomly sampled variable subset for training. 
We calculate the mean and standard deviation of the 10 models as the reported results. 
Additionally, the forecasting horizon length $Q$ is set as 12 and the lookback window $L$ is 12, unless specified otherwise.
Regarding the segmentation of the dataset, 70\% of the samples are for training, 10\% is for validation, and 20\% is for testing.

\subsection{Overall Performance (RQ1)} \label{overall-perforemance}
Table \ref{tab: overall performance} demonstrates the overall performance of four backbone models in time series forecasting on subset variables, along with their performances after incorporating the TOI-VSF. 
As detailed in Table~\ref{tab: overall performance}, irrespective of variations in datasets or the specific forecasting models applied, the inclusion of TOI-VSF leads to a consistent enhancement in forecasting accuracy.
Our model demonstrates significant improvements, with improvements exceeding $10\%$ in both mean absolute error (MAE) and root mean square error (RMSE) for most instances. It is worth noting that in some cases the improvement is more significant, exceeding $40\%$. This demonstrates the robustness of the TOI-VSF ensemble in improving the predictive power of time series models across a variety of scenarios, like transportation, healthcare, etc.
The bottom row of the table summarizes these results, showing the average percentage improvement for the different forecasting models. 
This consolidated data emphasizes the significant impact of the TOI-VSF method on overall forecasting performance, affirming its effectiveness as a potent tool for improving the accuracy of variable subset forecasting.

\begin{table*}[!h]
\centering
\small  
\setlength\tabcolsep{0.8pt}
\caption{Ablation Study of Joint Learning. `w/o jl' means the self-supervised variable imputation model undergoes solely pre-training prior to the training of the forecasting model.}
\begin{tabular}{c|c|cc|cc|cc|cc|cc}
\toprule
    \multicolumn{2}{c|}{\multirow{2.7}{*}{Forecasting Model}} & \multicolumn{2}{c|}{METR-LA}    & \multicolumn{2}{c|}{TRAFFIC}     & \multicolumn{2}{c|}{SOLAR}     & \multicolumn{2}{c}{ECG5000}   & \multicolumn{2}{c}{ETTH1}\\
\cmidrule{3-12}
\multicolumn{2}{c|}{}                                  & MAE           & RMSE           & MAE            & RMSE           & MAE           & RMSE          & MAE           & RMSE          & MAE           & RMSE\\
\midrule
\multirow{2.3}{*}{ASTGCN}      & w/o jl     & 9.26(1.57)          & 12.86(2.21)          & 32.24(2.51)          & 60.13(3.25)          & 9.74(2.21)          & 11.62(3.14)         & 6.48(1.06)          & 9.09(1.26)          & 4.92(1.86)          & 6.33(2.23)\\
                             & joint learning         & \textbf{4.34(0.36)} & \textbf{8.13(0.58)}  & \textbf{18.90(0.91)} & \textbf{39.69(2.11)} & \textbf{3.14(0.30)} & \textbf{4.74(0.57)} & \textbf{3.18(0.50)} & \textbf{5.40(0.96)} & \textbf{1.98(1.16)} & \textbf{3.23(2.07)}\\
\midrule
\multirow{2.3}{*}{MSTGCN}      & w/o jl     & 9.02(0.98)          & 11.98(1.05)          & 31.55(2.67)          & 58.22(3.17)          & 9.82(2.16)          & 11.03(3.01)         & 6.38(1.01)          & 10.03(1.92)         & 4.97(1.28)          & 6.27(2.18)\\
                             & joint learning         & \textbf{3.91(0.24)} & \textbf{7.66(0.44)}  & \textbf{17.36(0.84)} & \textbf{37.65(2.20)} & \textbf{2.56(0.25)} & \textbf{4.04(0.54)} & \textbf{3.18(0.50)} & \textbf{5.41(0.96)} & \textbf{1.94(1.08)} & \textbf{3.24(1.95)}\\
 \midrule
\multirow{2.3}{*}{MTGNN}       & w/o jl     & 8.75(1.64)          & 11.62(2.04)          & 28.17(3.12)          & 40.62(3.79)          & 8.89(1.24)          & 10.85(1.67)         & 6.12(1.53)          & 9.52(2.20)          & 4.29(1.27)          & 7.81(1.74)\\
                             & joint learning         & \textbf{3.28(0.21)} & \textbf{6.51(0.38)}  & \textbf{11.40(0.61)} & \textbf{27.35(2.02)} & \textbf{2.26(0.23)} & \textbf{3.53(0.47)} & \textbf{3.21(0.49)} & \textbf{5.51(0.93)} & \textbf{1.72(0.95)} & \textbf{2.88(1.68)}\\
\midrule
\multirow{2.3}{*}{TGCN}        & w/o jl     & 12.38(1.22)         & 18.57(2.56)          & 46.72(2.82)          & 70.61(3.23)          & 10.30(1.61)         & 15.39(2.44)         & 8.21(2.53)          & 12.39(3.40)         & 6.23(1.98)          & 9.31(2.90)\\
                             & joint learning         & \textbf{8.46(0.94)} & \textbf{14.14(1.28)} & \textbf{28.82(1.46)} & \textbf{51.14(2.75)} & \textbf{4.43(0.78)} & \textbf{7.15(1.57)} & \textbf{6.13(1.42)} & \textbf{9.63(2.36)} & \textbf{3.60(2.03)} & \textbf{5.32(2.46)}\\
\bottomrule
\end{tabular}
\label{tab:nojoint-learning}
\end{table*}

\begin{figure*}[!ht]
    \centering
    \hspace{-2mm}
    \subfloat[ASTGCN]{
        \includegraphics[width=0.235\linewidth]{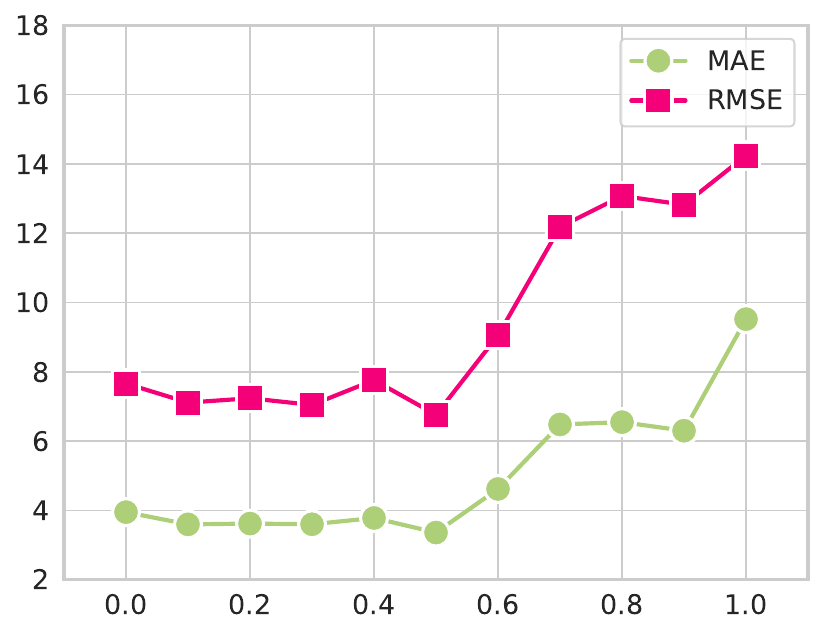}
        }
        \hspace{-2mm}
    \subfloat[MSTGCN]{
        \includegraphics[width=0.235\linewidth]{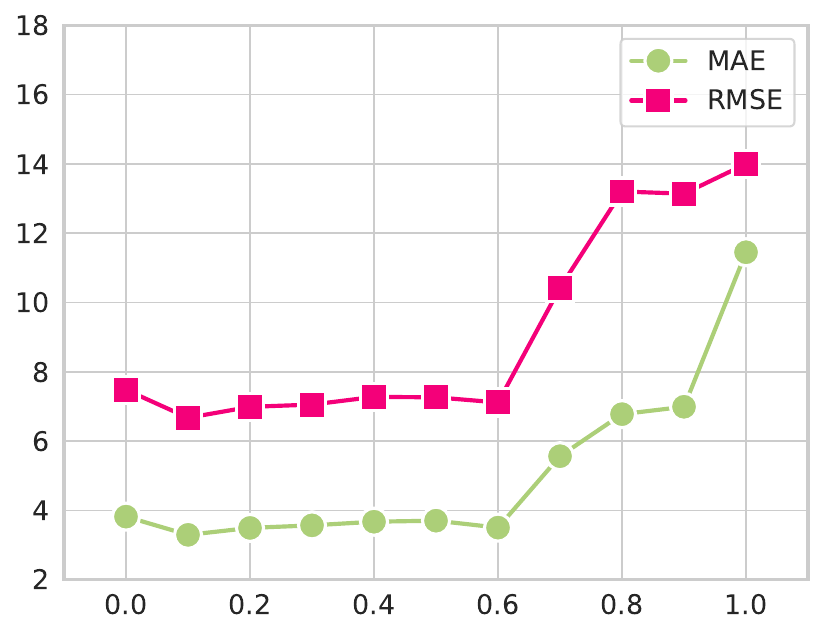}
        }
        \hspace{-2mm}
    \subfloat[MTGNN]{
        \includegraphics[width=0.235\linewidth]{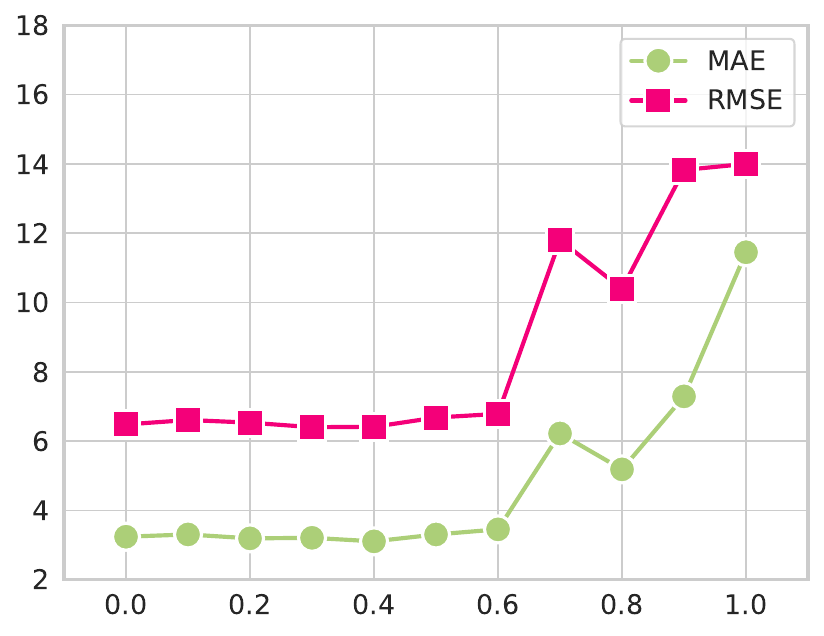}
        }
        \hspace{-2mm}
    \subfloat[TGCN]{
        \includegraphics[width=0.235\linewidth]{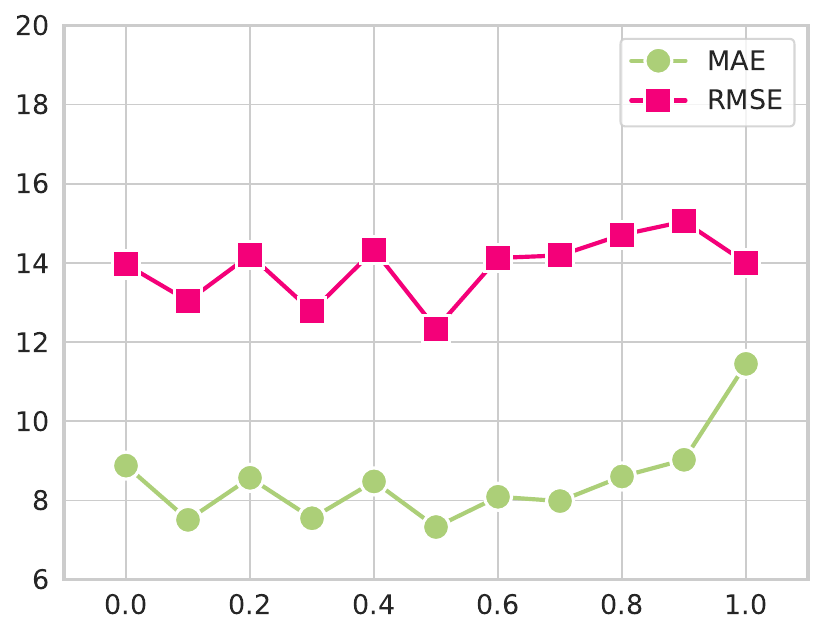}
        }
        \hspace{-2mm}
    \caption{Different Weights of Self-supervised Learning Module Loss on METR-LA. The horizontal axis is $\alpha$, which means the proportion of self-supervised learning loss to the overall loss.}

    \label{fig:weight-metr}
\end{figure*}
\begin{figure*}[!h]
    \centering
    \hspace{-2mm}
    \subfloat[ASTGCN]{
        \includegraphics[width=0.235\linewidth]{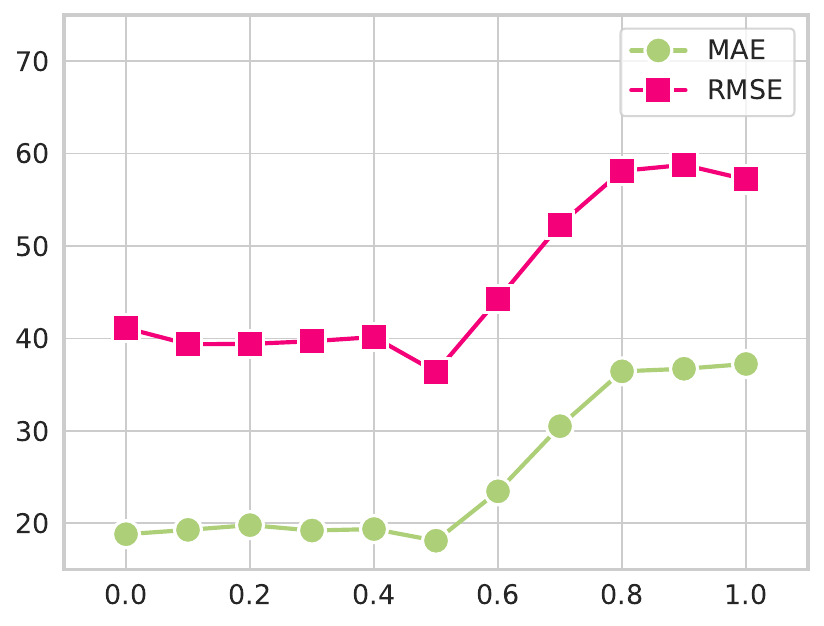}
        }
        \hspace{-2mm}
    \subfloat[MSTGCN]{
        \includegraphics[width=0.235\linewidth]{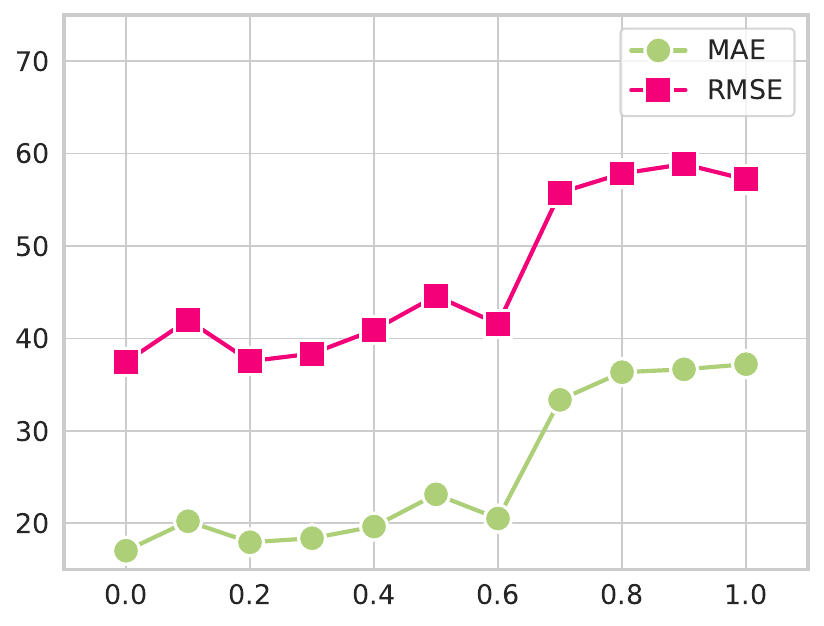}
        }
        \hspace{-2mm}
    \subfloat[MTGNN]{
        \includegraphics[width=0.235\linewidth]{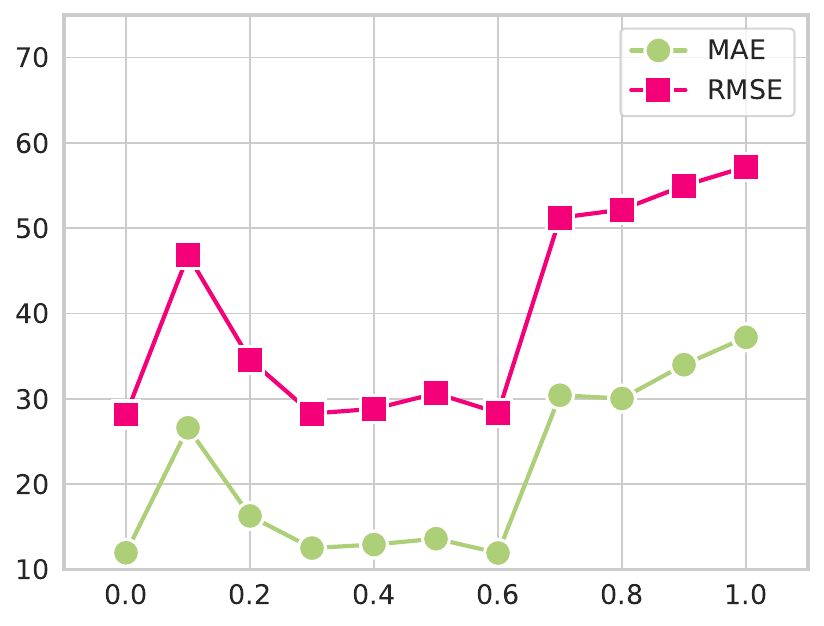}
        }
        \hspace{-2mm}
    \subfloat[TGCN]{
        \includegraphics[width=0.235\linewidth]{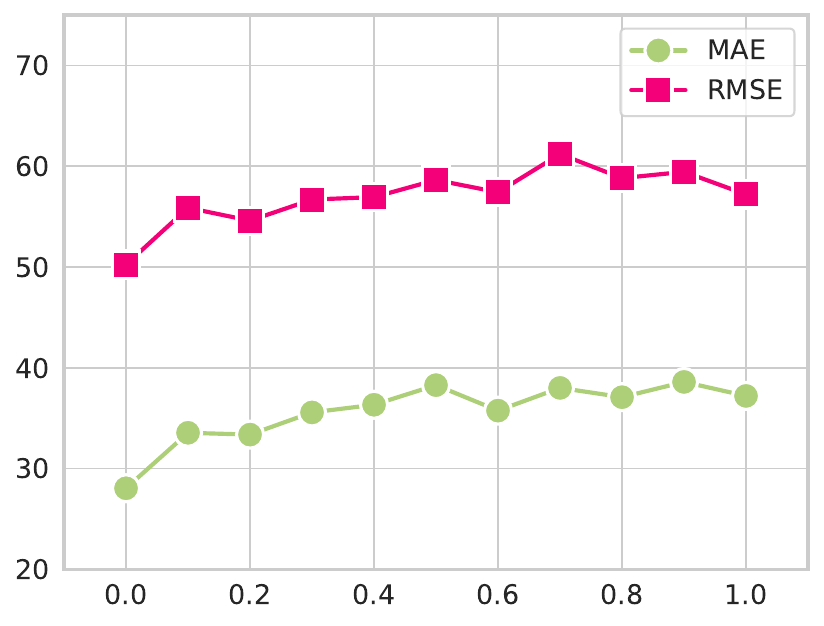}
        }
        \hspace{-2mm}
    \caption{Different Weights of Self-supervised Learning Module Loss on TRAFFIC. The horizontal axis is $\alpha$, which means the proportion of self-supervised learning loss to the overall loss.}

    \label{fig:weight-traffic}
\end{figure*}

\subsection{Comparison with \textit{Oracle} Setting (RQ2)}
Table~\ref{tab:oracle-ours} illustrates the comparative performance between our proposed model and the \textit{Oracle} setting.
Notably, our model consistently outperforms the \textit{Oracle} setting by an average margin of over $5\%$.
A potential reason for this significant improvement in forecasting performance is the consideration of joint learning between the self-supervised variable imputation model and the forecasting model. 
This joint learning approach allows for a synergistic interaction where the forecasting model actively informs and guides the imputation model. This guidance ensures that the missing data are filled in a manner that is most conducive to enhancing the forecast accuracy. Such a strategy is particularly effective in addressing and mitigating common data issues that typically impair forecasting, such as biases, redundancies, and harmful inter-variable relationships.
Furthermore, this process not only improves the robustness of the forecasting outcomes but also enhances the overall reliability of the predictive analytics framework. 
By integrating the forecasting and imputation processes, our model effectively turns potential data weaknesses into strengths, leveraging every piece of available data towards more accurate and reliable predictions.
A deeper insight into how this integration impacts forecasting performance is in Section~\ref{joint-learning}.

\begin{figure*}[!ht]
    \centering
    \hspace{-2mm}
    \subfloat[ASTGCN]{
        \includegraphics[width=0.235\linewidth]{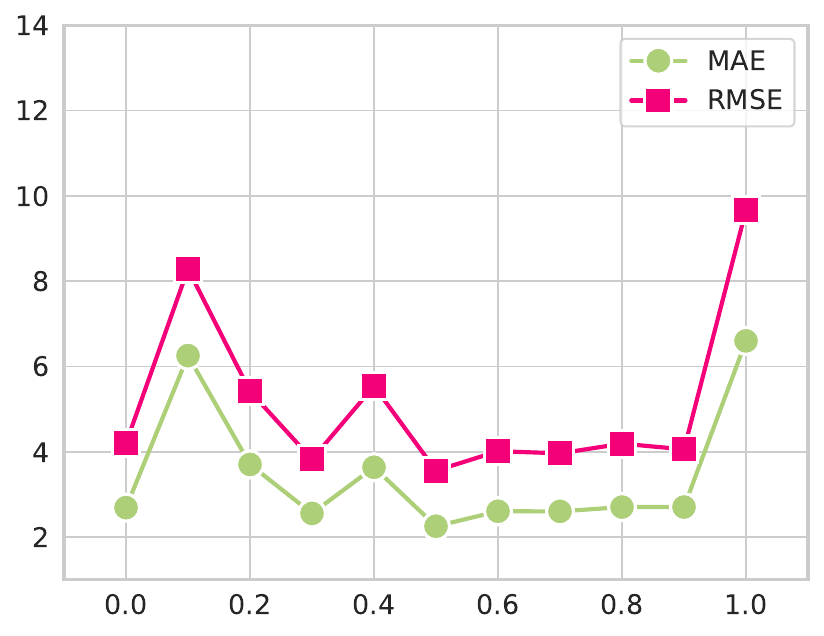}
        }
        \hspace{-2mm}
    \subfloat[MSTGCN]{
        \includegraphics[width=0.235\linewidth]{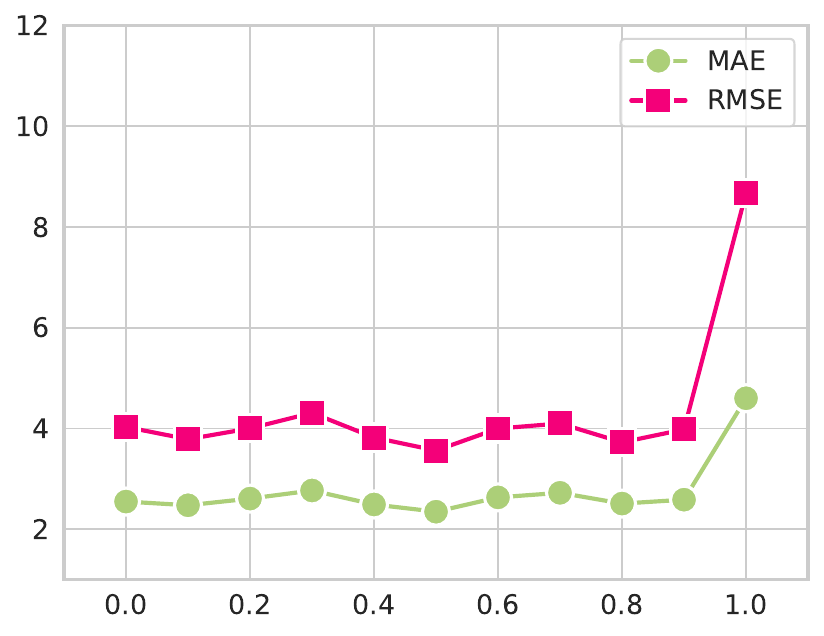}
        }
        \hspace{-2mm}
    \subfloat[MTGNN]{
        \includegraphics[width=0.235\linewidth]{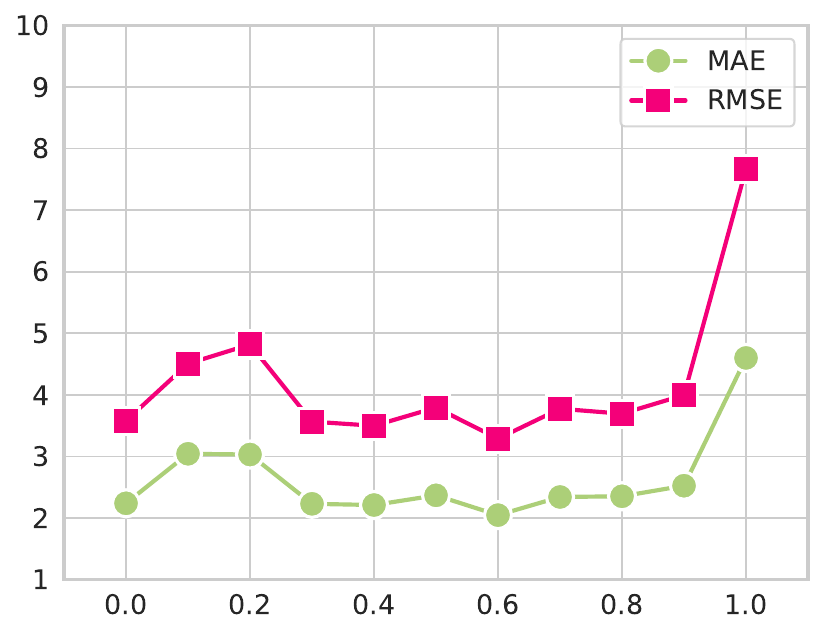}
        }
        \hspace{-2mm}
    \subfloat[TGCN]{
        \includegraphics[width=0.235\linewidth]{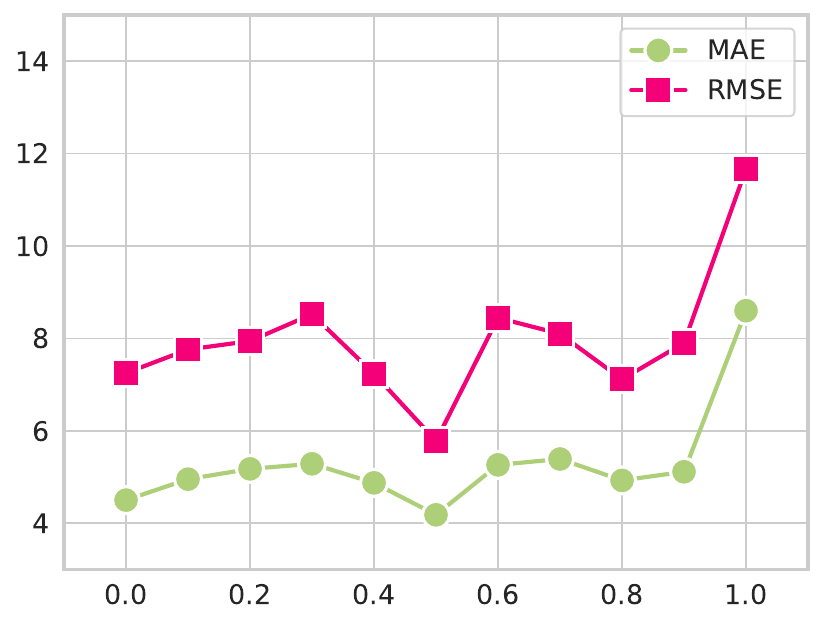}
        }
        \hspace{-2mm}
    \caption{Different Weights of Self-supervised Learning Module Loss on SOLAR. The horizontal axis is $\alpha$, which means the proportion of self-supervised learning loss to the overall loss.}

    \label{fig:weight-solar}
\end{figure*}
\begin{figure*}[!t]
    \centering
    \hspace{-2mm}
    \subfloat[ASTGCN]{
    \includegraphics[width=0.235\linewidth]{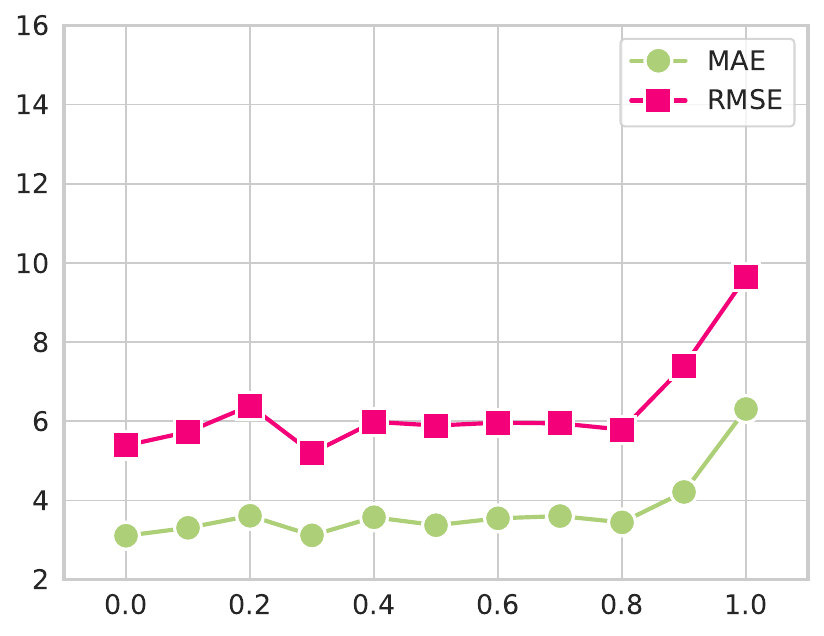}
        }
        \hspace{-2mm}
    \subfloat[MSTGCN]{
    \includegraphics[width=0.235\linewidth]{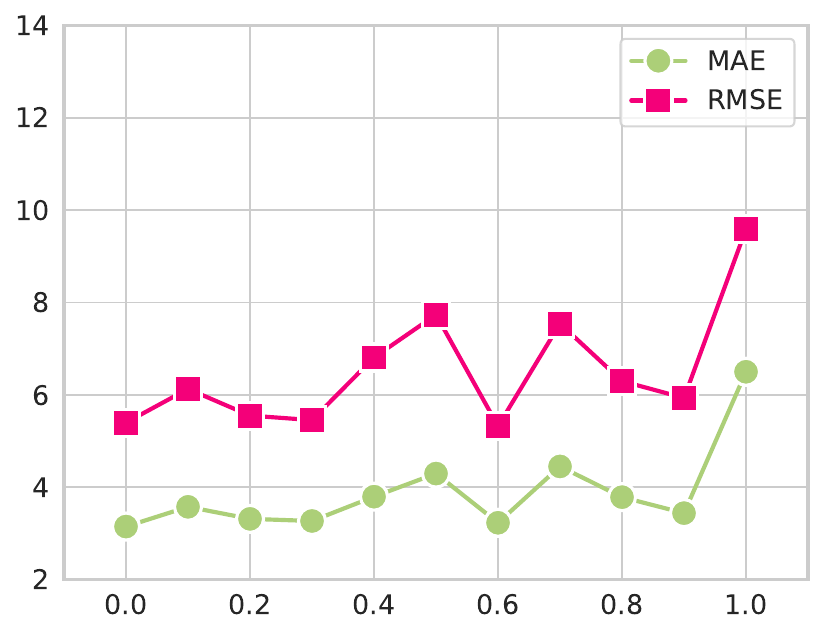}
        }
        \hspace{-2mm}
    \subfloat[MTGNN]{
    \includegraphics[width=0.235\linewidth]{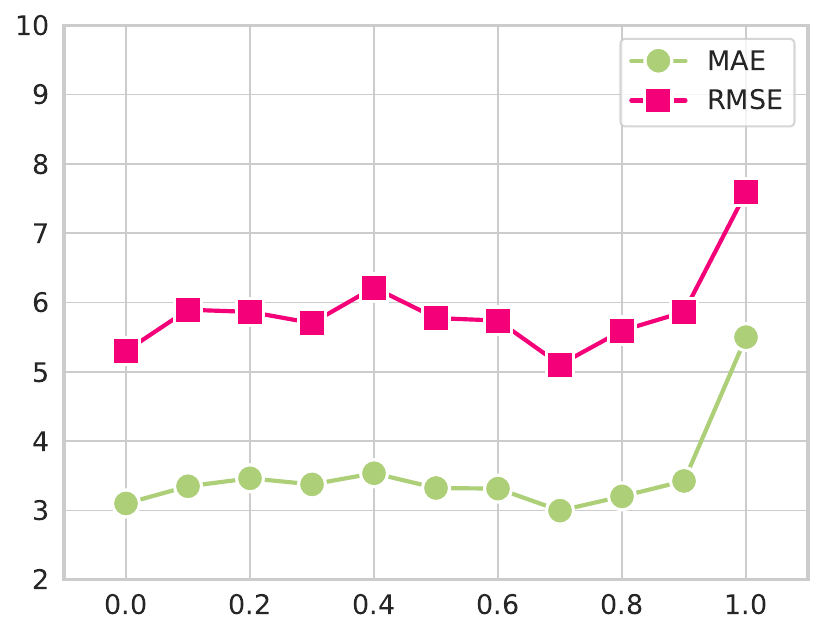}
        }
        \hspace{-2mm}
    \subfloat[TGCN]{
    \includegraphics[width=0.235\linewidth]{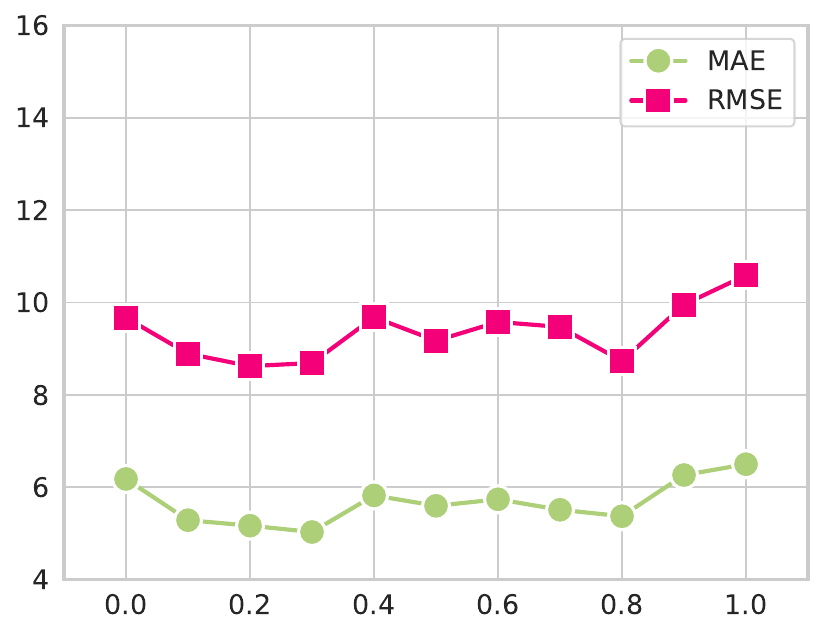}
        }
        \hspace{-2mm}
    \caption{Different Weights of Self-supervised Learning Module Loss on ECG5000. The horizontal axis is $\alpha$, which means the proportion of self-supervised learning loss to the overall loss.}

    \label{fig:weight-ecg}
\end{figure*}
\begin{figure*}[!ht]
    \centering
    \hspace{-2mm}
    \subfloat[ASTGCN]{
    \includegraphics[width=0.235\linewidth]{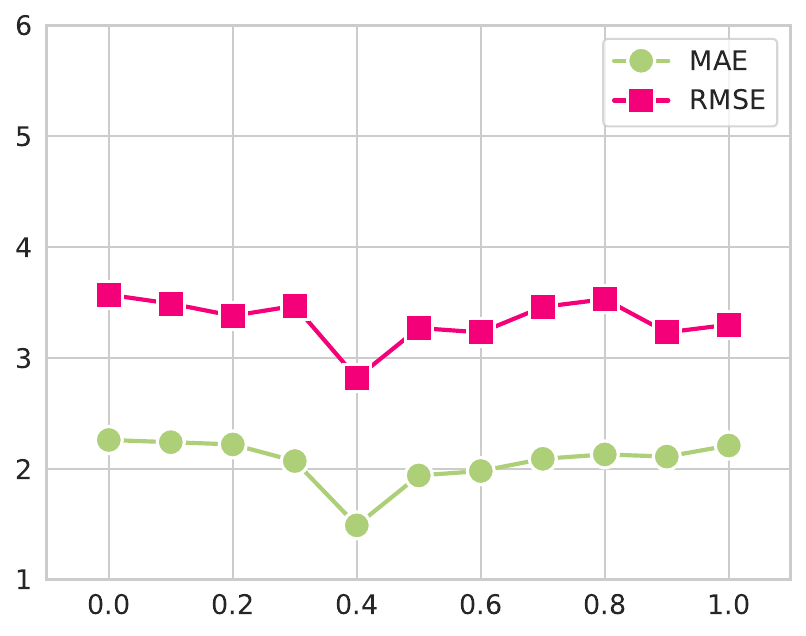}
        }
        \hspace{-2mm}
    \subfloat[MSTGCN]{
    \includegraphics[width=0.235\linewidth]{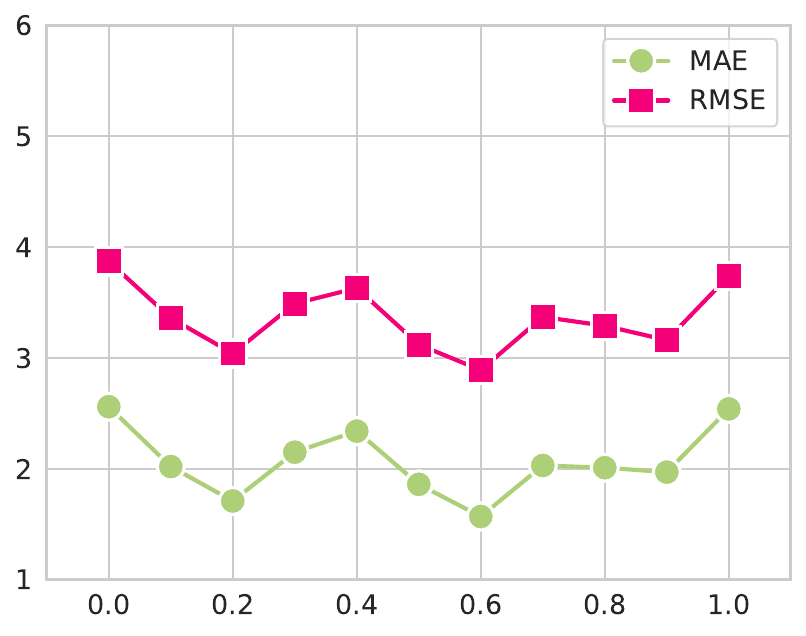}
        }
        \hspace{-2mm}
    \subfloat[MTGNN]{
    \includegraphics[width=0.235\linewidth]{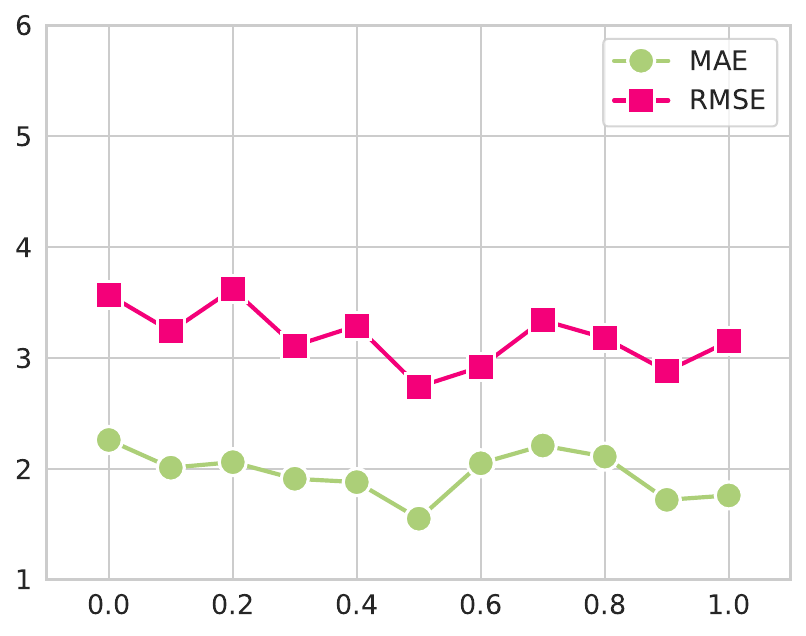}
        }
        \hspace{-2mm}
    \subfloat[TGCN]{
    \includegraphics[width=0.235\linewidth]{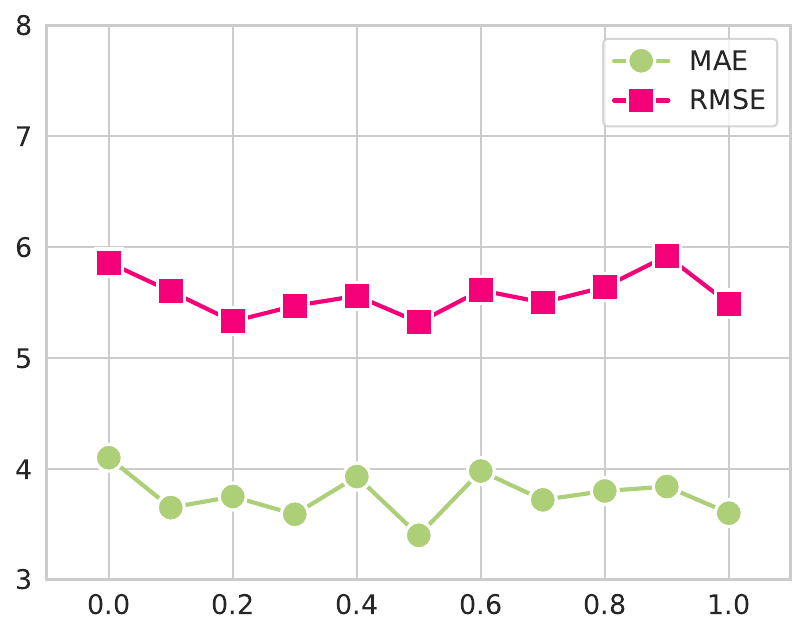}
        }
        \hspace{-2mm}
    \caption{Different Weights of Self-supervised Learning Module Loss on ETTH1. The horizontal axis is $\alpha$, which means the proportion of self-supervised learning loss to the overall loss.}

    \label{fig:weight-etth1}
\end{figure*}

\subsection{Analysis of Joint Learning (RQ3)}\label{joint-learning}
To evaluate the effectiveness of joint learning, we compare the performance with and without the utilization of joint learning in Table~\ref{tab:nojoint-learning}.
\begin{itemize}
\item\noindent \textbf{w/o jl (joint learning):} The self-supervised variable imputation model is pre-trained independently before the training of the forecasting model begins.
Post pre-training, it is integrated solely during the inference phase.
\end{itemize}

The results displayed in Table~\ref{tab:nojoint-learning} clearly show a huge drop in performance when joint learning is not utilized, highlighting its significant contribution to model effectiveness.
Generally speaking, for the \textit{oracle} setting, the input data is complete, and only the forecasting model is trained.
The \textit{Oracle} setting serves as a benchmark, providing an upper bound to the forecasting results.
While, through joint learning, our method TOI-VSF not only meets but exceeds the \textit{Oracle} setting. This superior performance suggests that the joint learning of both the self-supervised variable imputation model and the forecasting model allows for shared and dynamically updated parameters, leading to enhanced outcomes. 
At the same time, joint learning entails the simultaneous training of both tasks, integrating information from both the imputation and the forecasting phases. This dual-task approach enables the model to develop a more comprehensive understanding of the input data, capture shared features between tasks, and reduce the risk of overfitting.
The synergistic interaction between the variable imputation and forecasting tasks under joint learning further enhances the overall forecasting performance, demonstrating the benefit of our joint training approach.

\begin{figure}[!th]
\centering
\hspace{-8mm}
\subfloat[MAE]{
\includegraphics[width=0.49\linewidth]{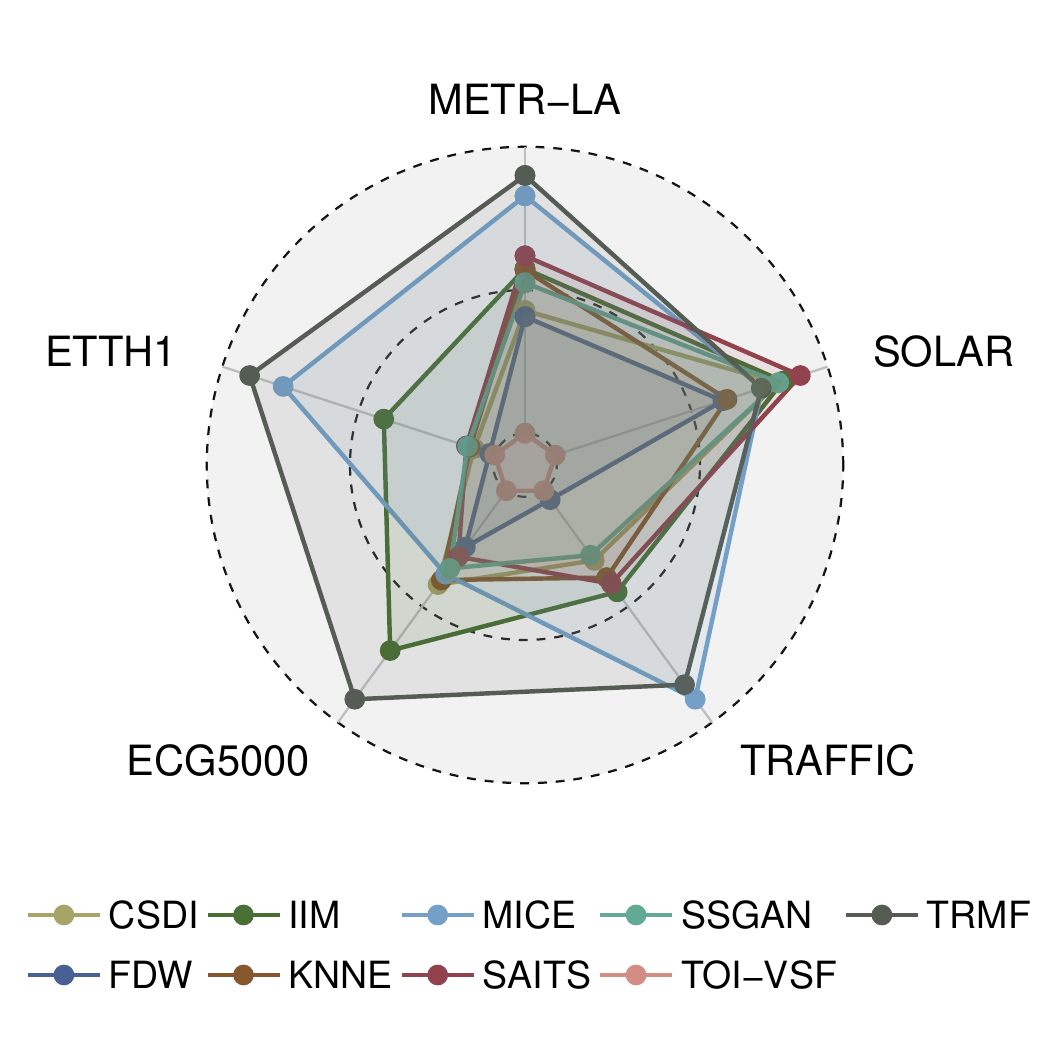}
}
\hspace{-5mm}
\subfloat[RMSE]{
\includegraphics[width=0.49\linewidth]{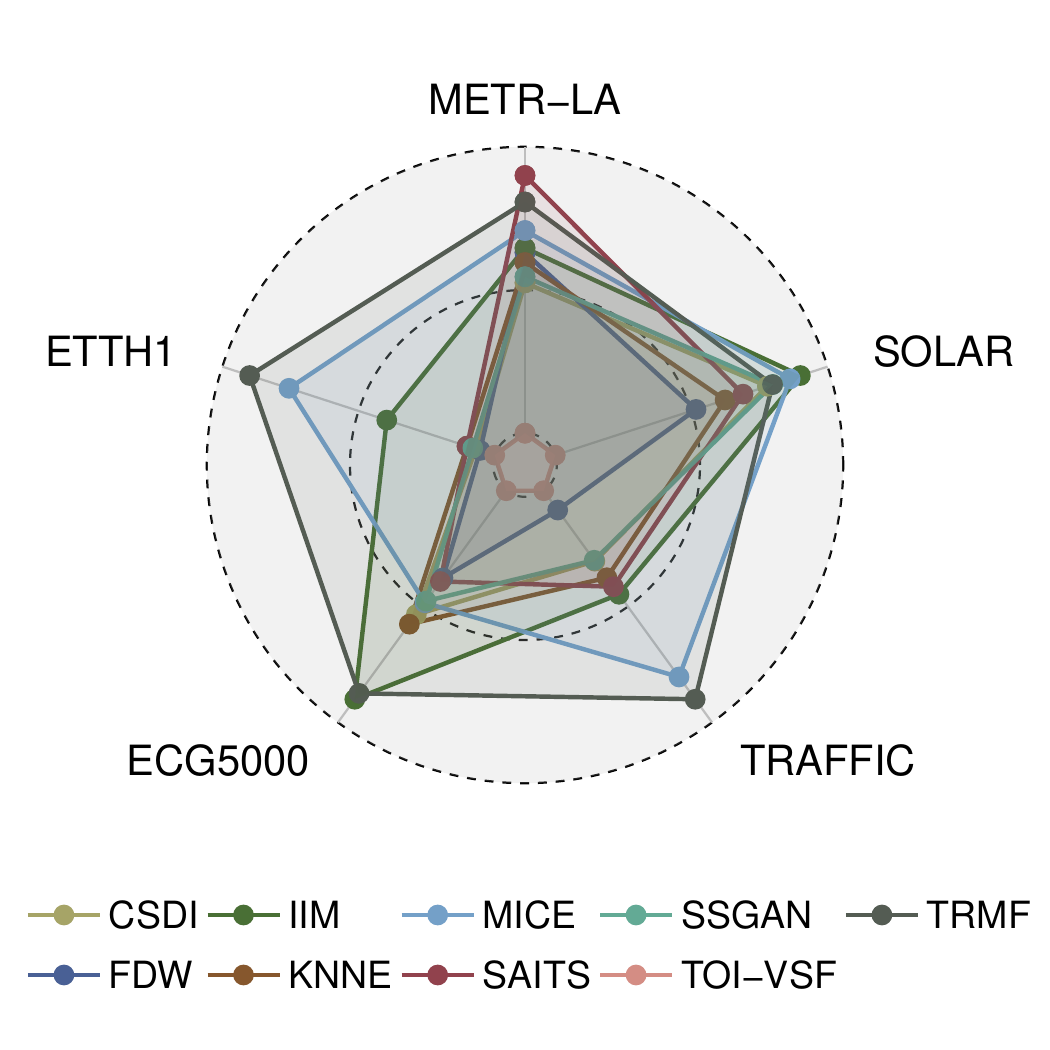}
}
\hspace{-7mm}
\caption{Performance of different imputation methods on VSF, utilizing the backbone ASTGCN. The lower the results, the better the performance.}
\label{fig: imputation-astgcn}
\end{figure}

\begin{figure}[!th]
\centering
\hspace{-8mm}
\subfloat[MAE]{
\includegraphics[width=0.49\linewidth]{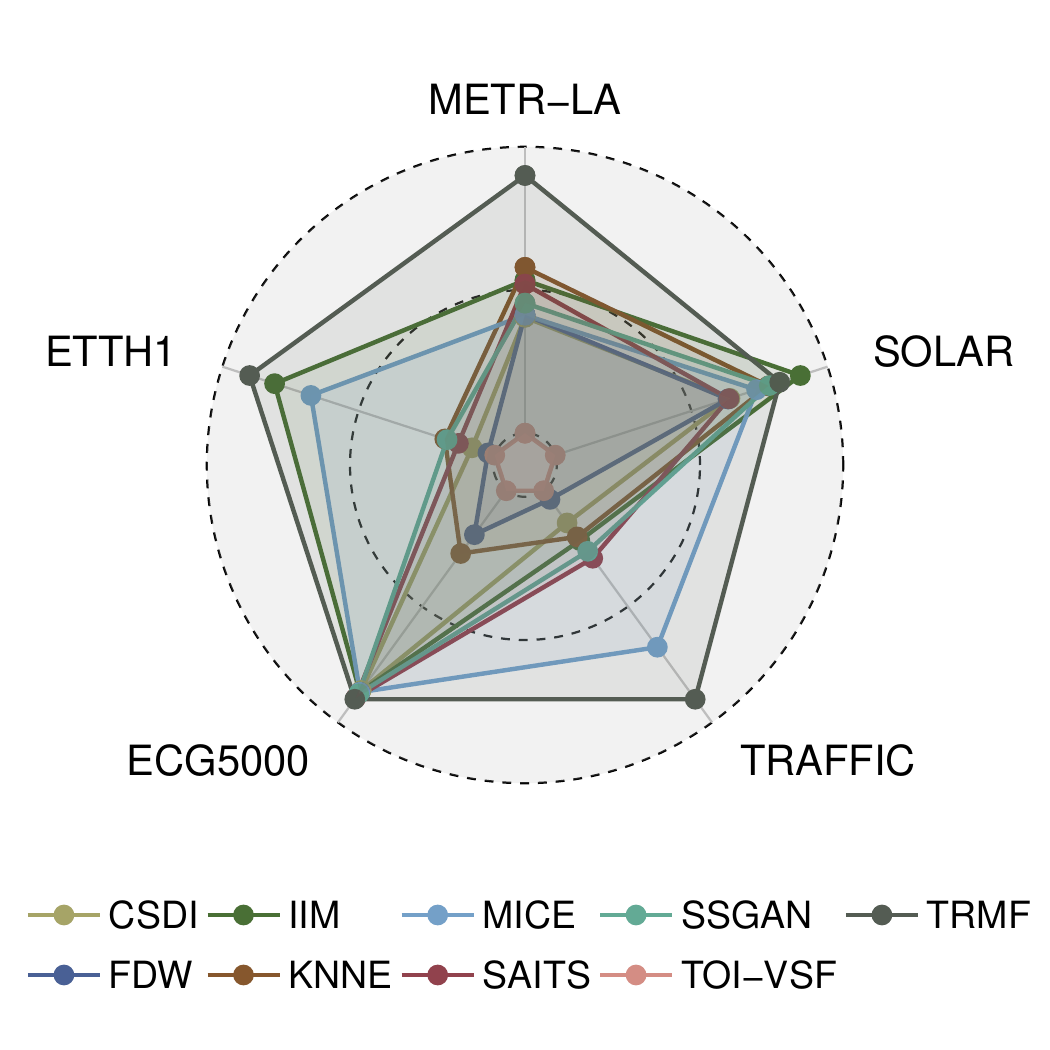}
}
\hspace{-5mm}
\subfloat[RMSE]{
\includegraphics[width=0.49\linewidth]{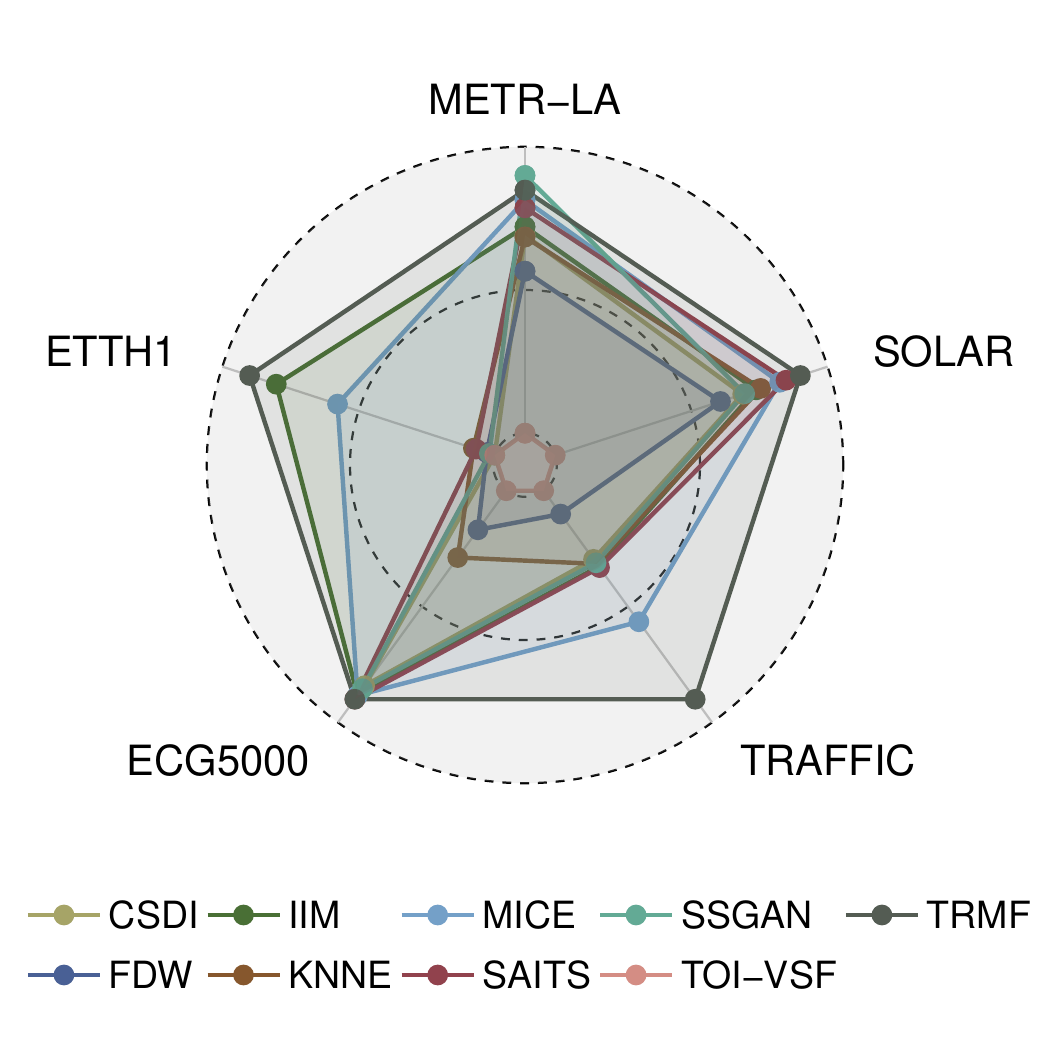}
}
\hspace{-7mm}
\caption{Performance of different imputation methods on VSF, utilizing the backbone MSTGCN. The lower the results, the better the performance.}
\label{fig: imputation-mstgcn}
\end{figure}

\begin{figure}[!th]
\centering
\hspace{-8mm}
\subfloat[MAE]{
\includegraphics[width=0.49\linewidth]{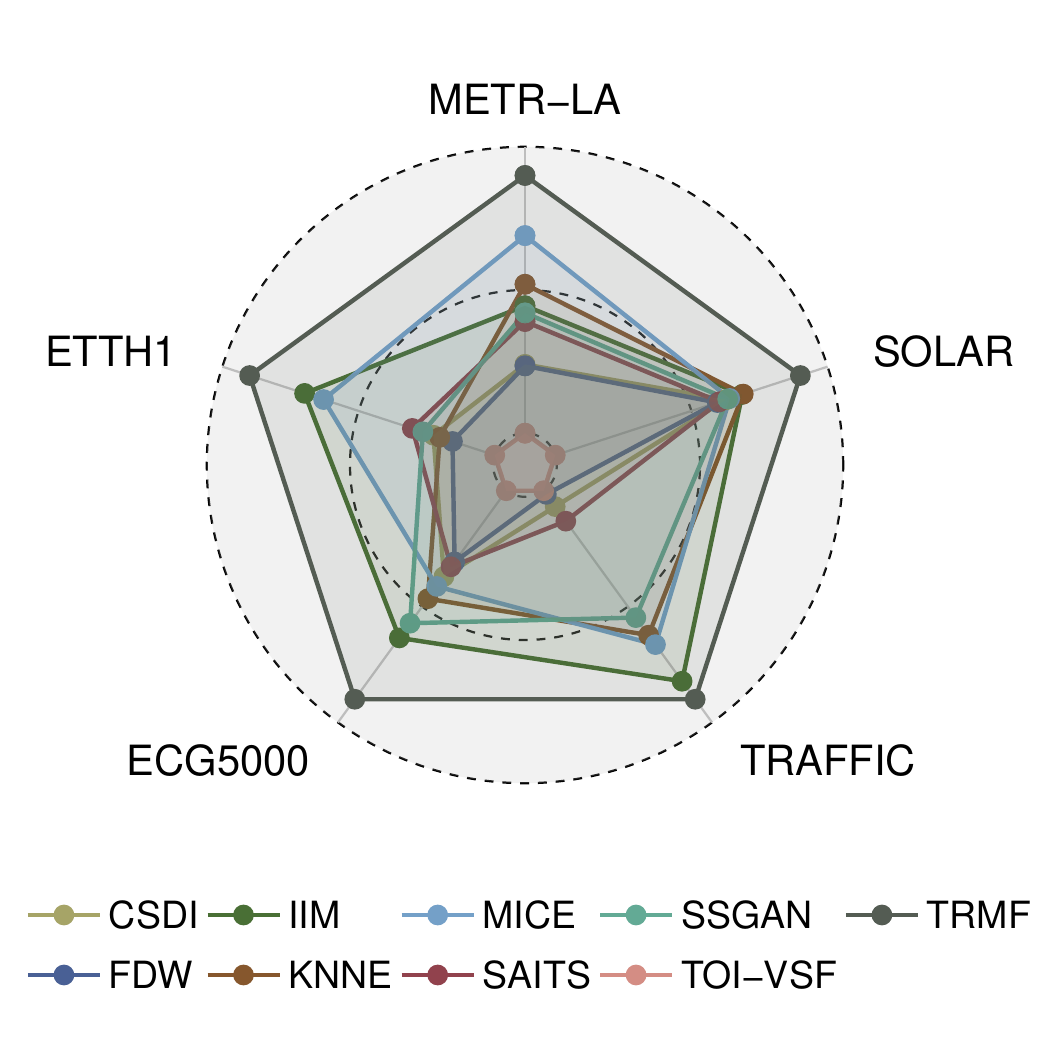}
}
\hspace{-5mm}
\subfloat[RMSE]{
\includegraphics[width=0.49\linewidth]{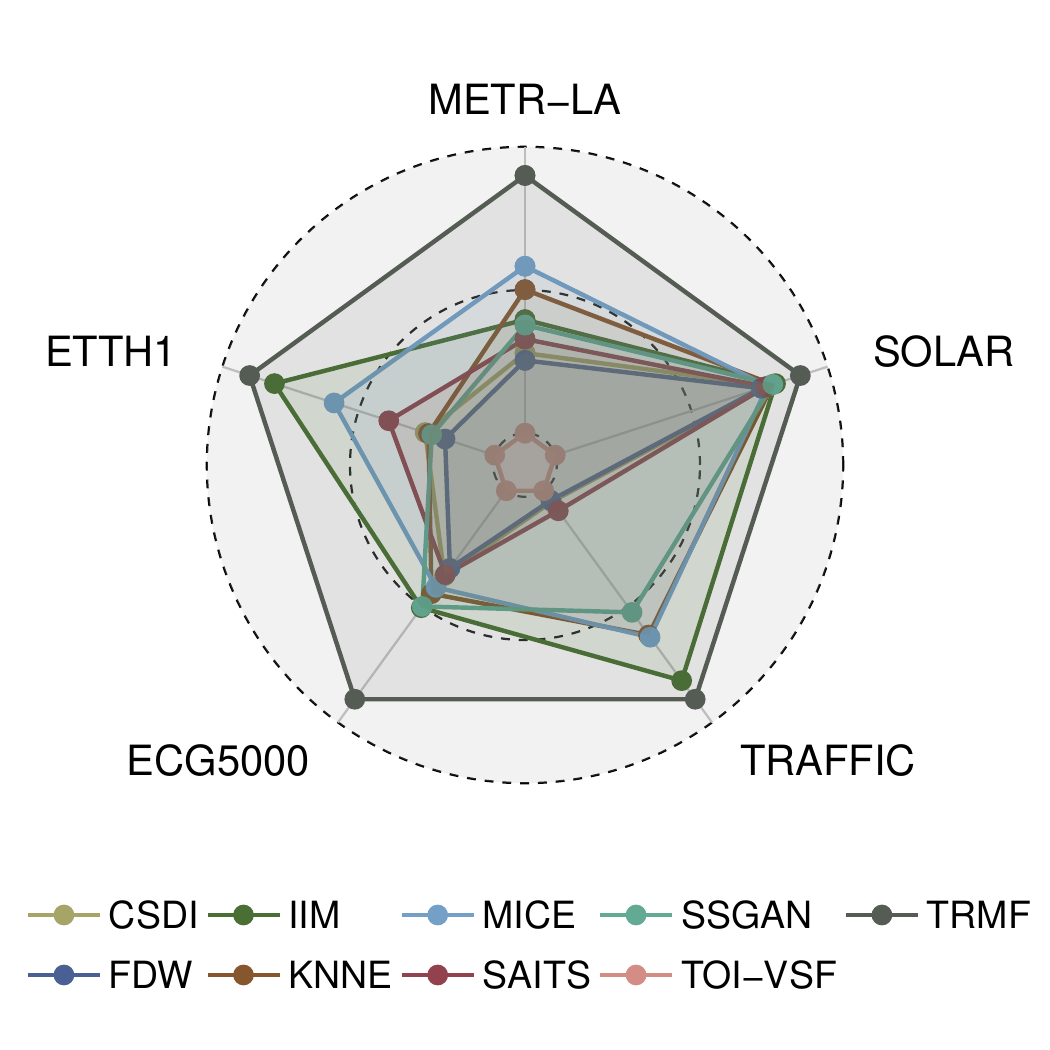}
}
\hspace{-7mm}
\caption{Performance of different imputation methods on VSF, utilizing the backbone MTGNN. The lower the results, the better the performance.}
\label{fig: imputation-mtgnn}
\end{figure}

\begin{figure}[!th]
\centering
\hspace{-8mm}
\subfloat[MAE]{
\includegraphics[width=0.49\linewidth]{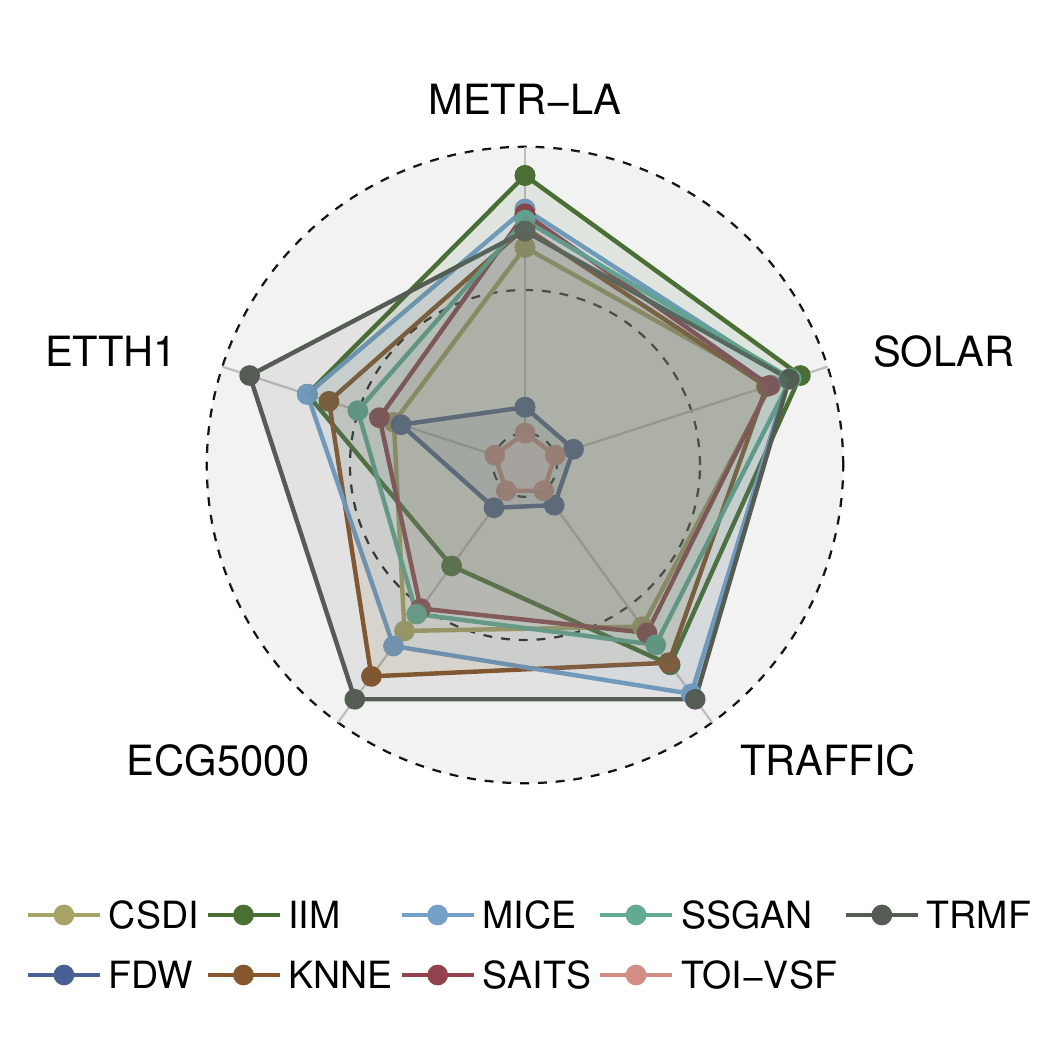}
}
\hspace{-5mm}
\subfloat[RMSE]{
\includegraphics[width=0.49\linewidth]{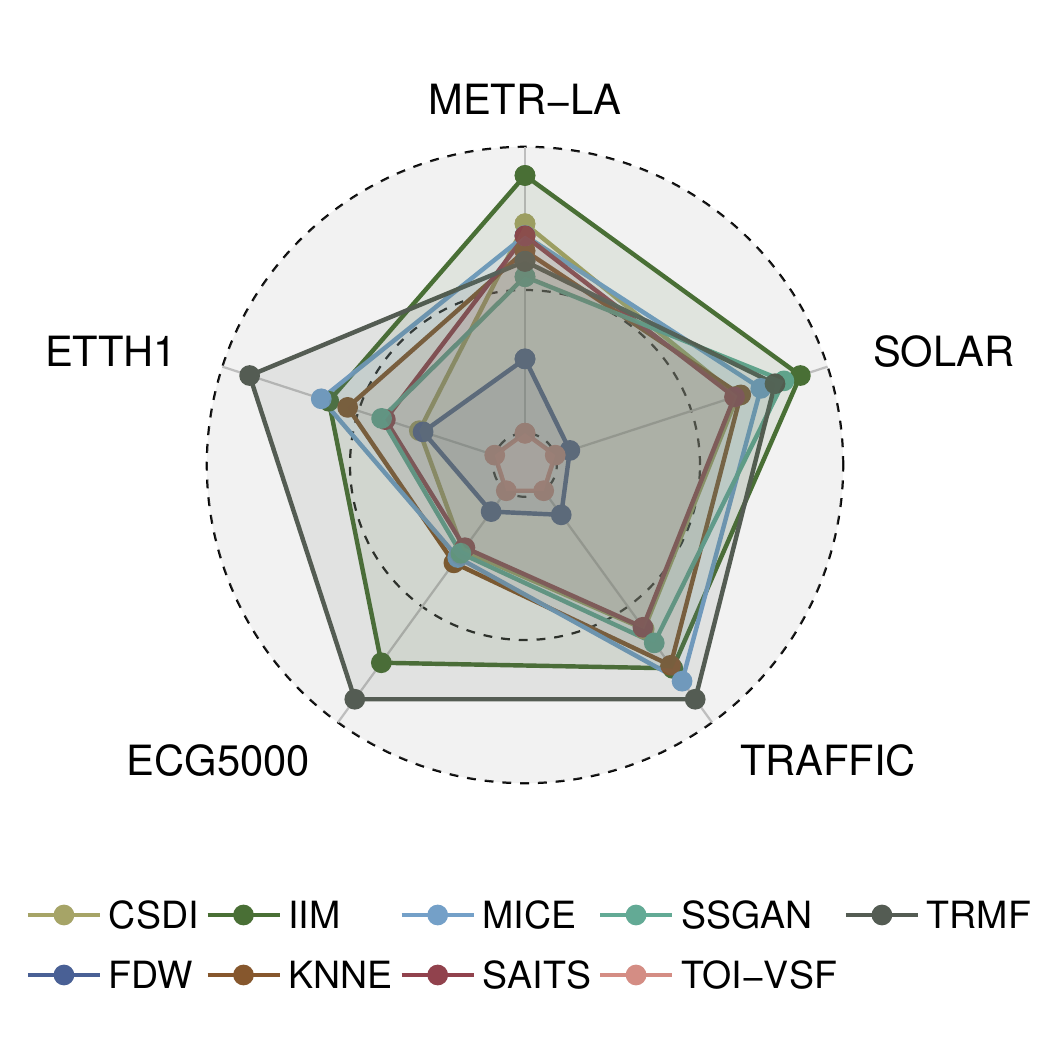}
}
\hspace{-7mm}
\caption{Performance of different imputation methods on VSF, utilizing the backbone TGCN. The lower the results, the better the performance.}
\label{fig: imputation-tgcn}
\end{figure}

\begin{figure}[!th]
\centering
\hspace{-8mm}
\subfloat[MAE]{
\includegraphics[width=0.49\linewidth]{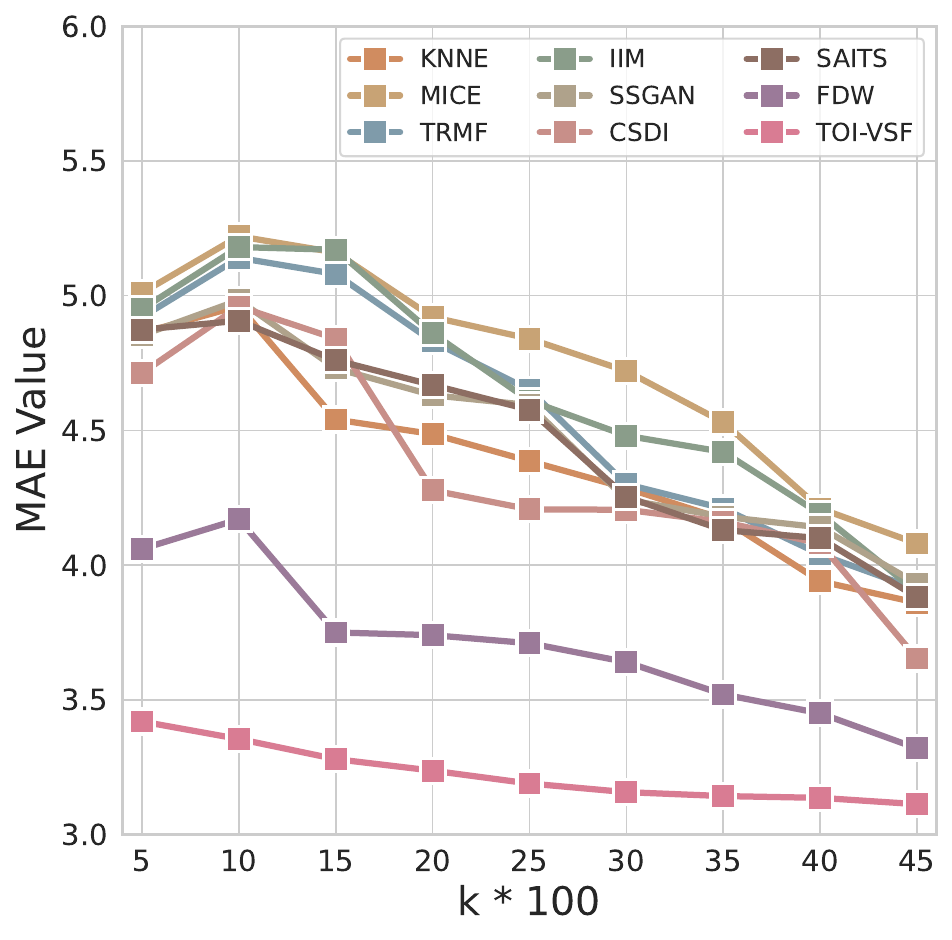}
}
\hspace{-2mm}
\subfloat[RMSE]{
\includegraphics[width=0.49\linewidth]{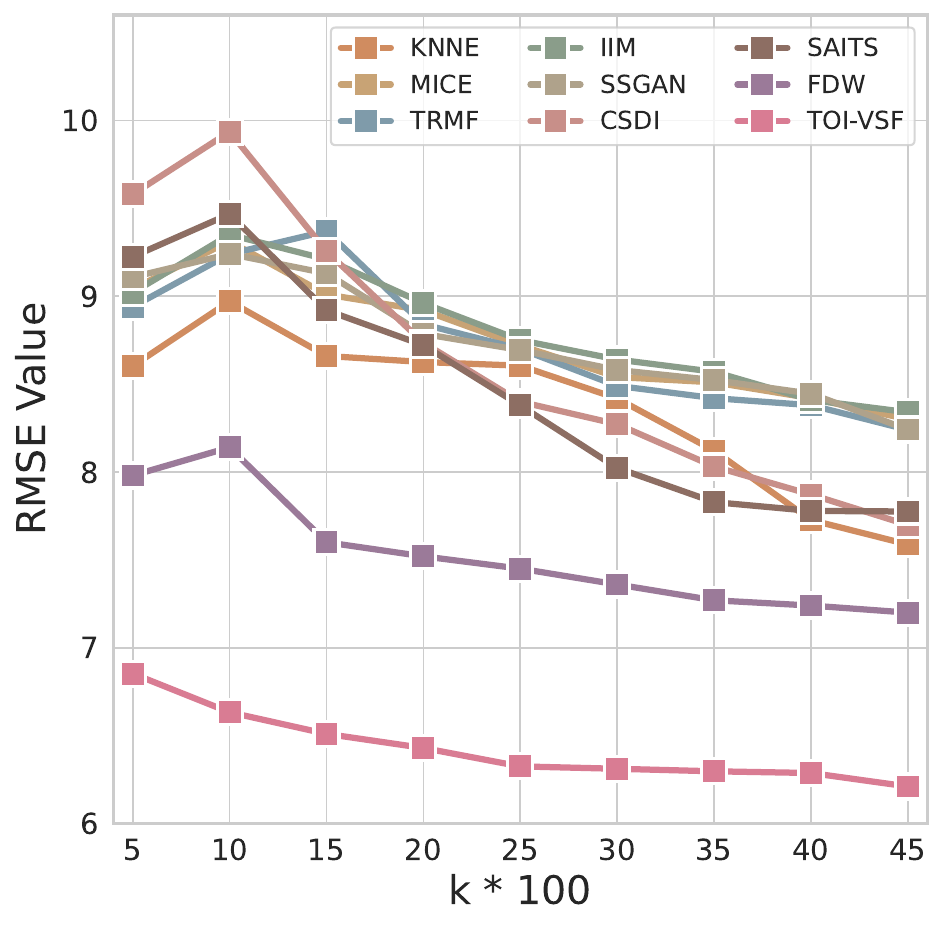}
}
\hspace{-7mm}
\caption{Performance of different $k$ values on VSF, utilizing the backbone MTGNN on the METR-LA dataset. The lower the results, the better the performance.}
\label{fig: k-metrla}
\end{figure}
\begin{figure}[!th]
\centering
\hspace{-8mm}
\subfloat[MAE]{
\includegraphics[width=0.49\linewidth]{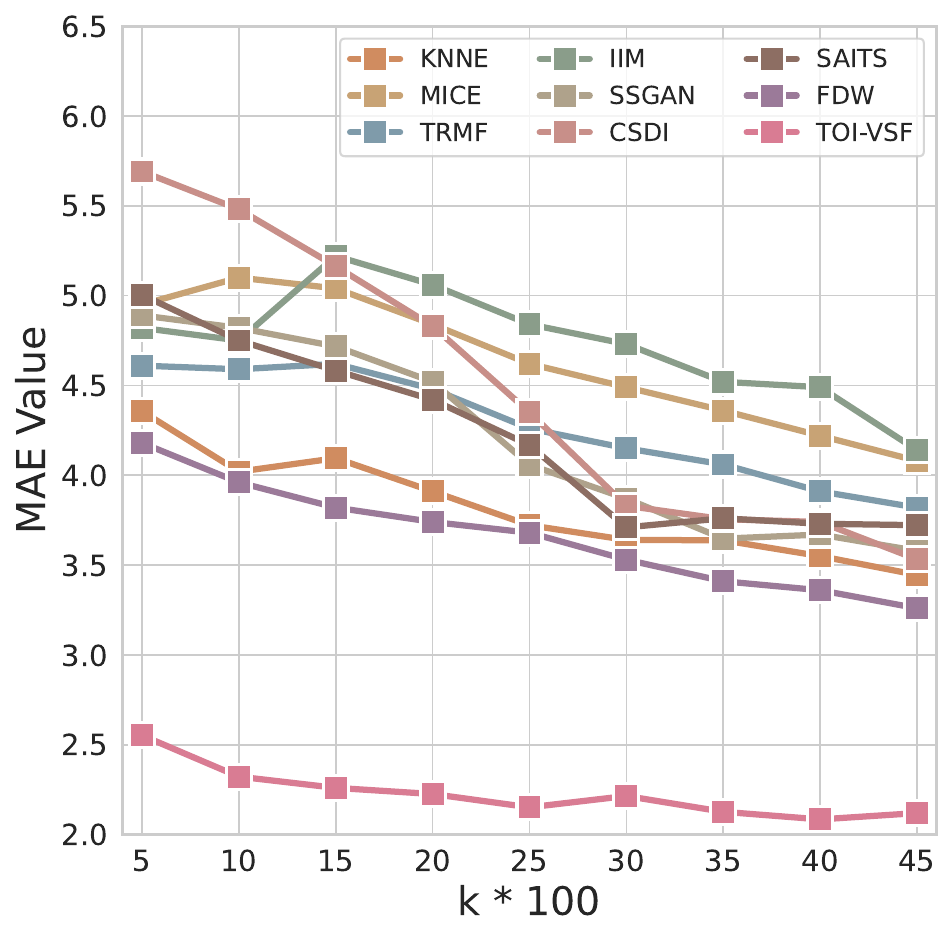}
}
\hspace{-2mm}
\subfloat[RMSE]{
\includegraphics[width=0.49\linewidth]{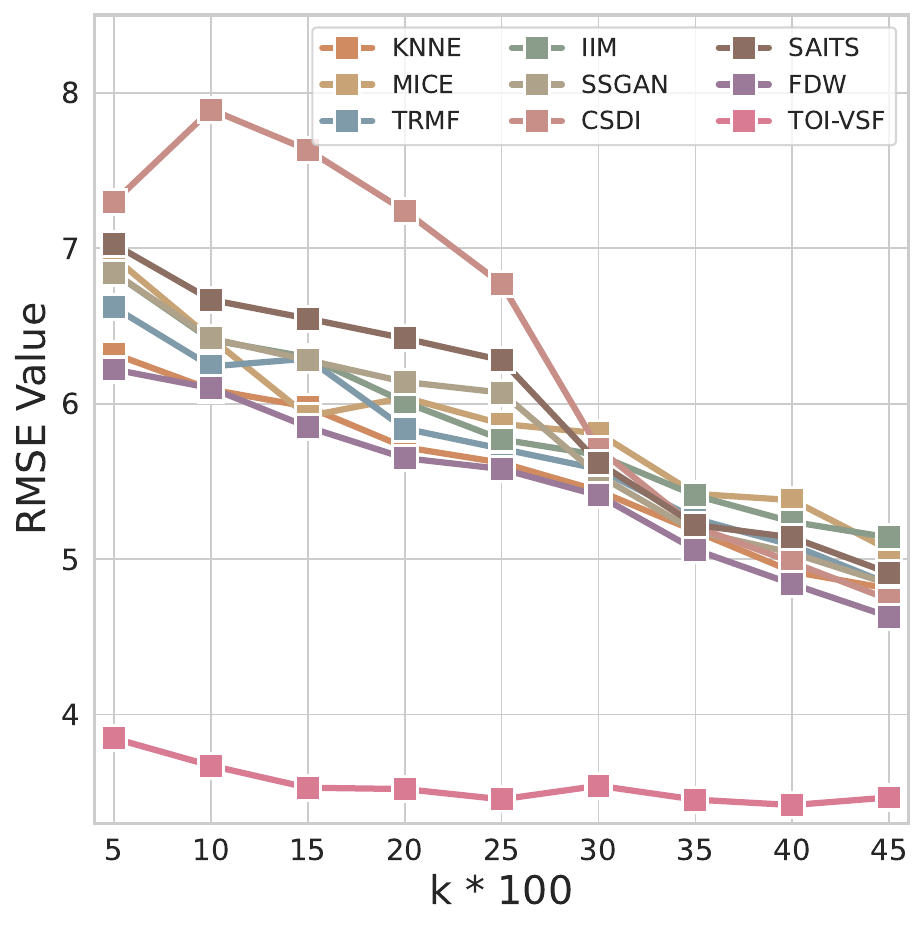}
}
\hspace{-7mm}
\caption{Performance of different $k$ values on VSF, utilizing the backbone MTGNN on the SOLAR dataset. The lower the results, the better the performance.}
\label{fig: k-solar}
\end{figure}
\begin{figure}[!th]
\centering
\hspace{-8mm}
\subfloat[MAE]{
\includegraphics[width=0.49\linewidth]{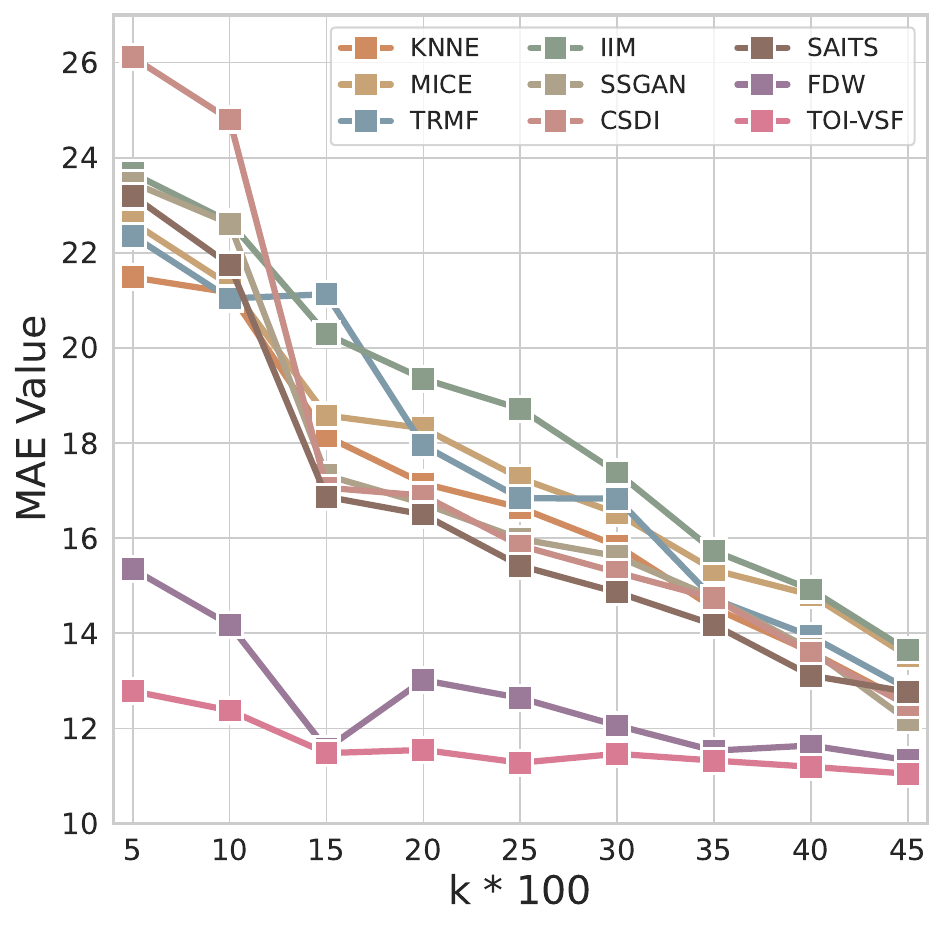}
}
\hspace{-2mm}
\subfloat[RMSE]{
\includegraphics[width=0.49\linewidth]{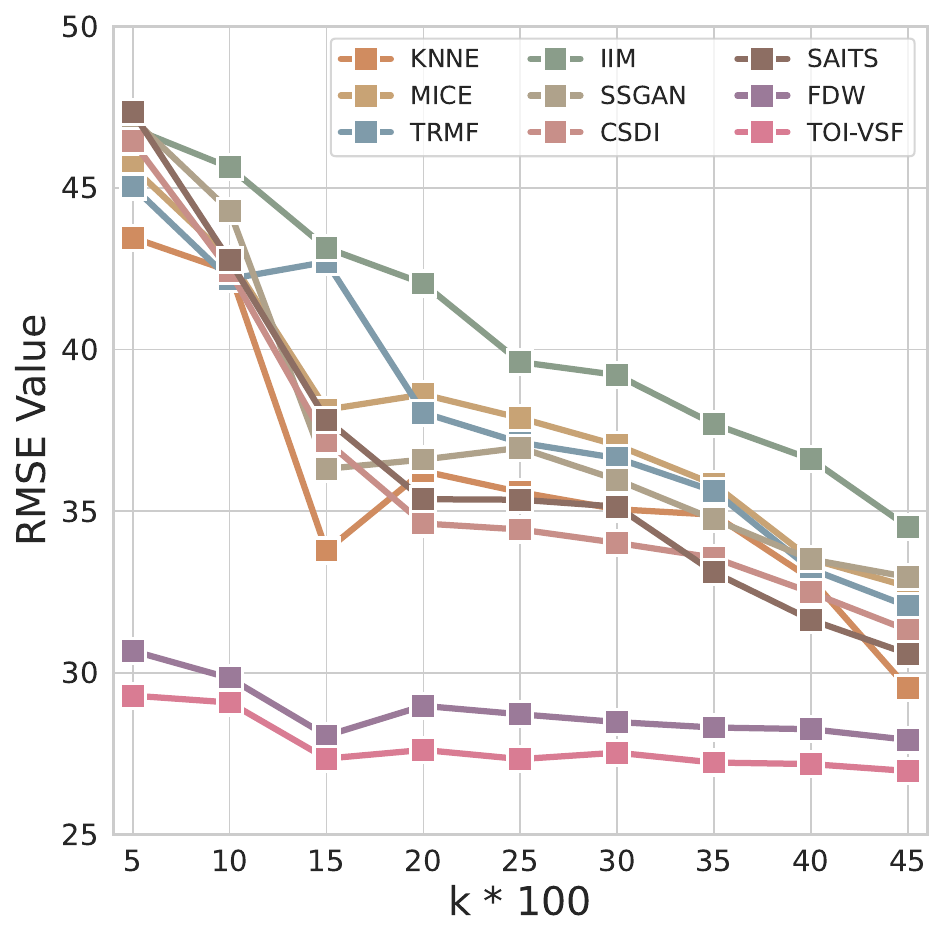}
}
\hspace{-7mm}
\caption{Performance of different $k$ values on VSF, utilizing the backbone MTGNN on the TRAFFIC dataset. The lower the results, the better the performance.}
\label{fig: k-traffic}
\end{figure}
\begin{figure}[!th]
\centering
\hspace{-8mm}
\subfloat[MAE]{
\includegraphics[width=0.49\linewidth]{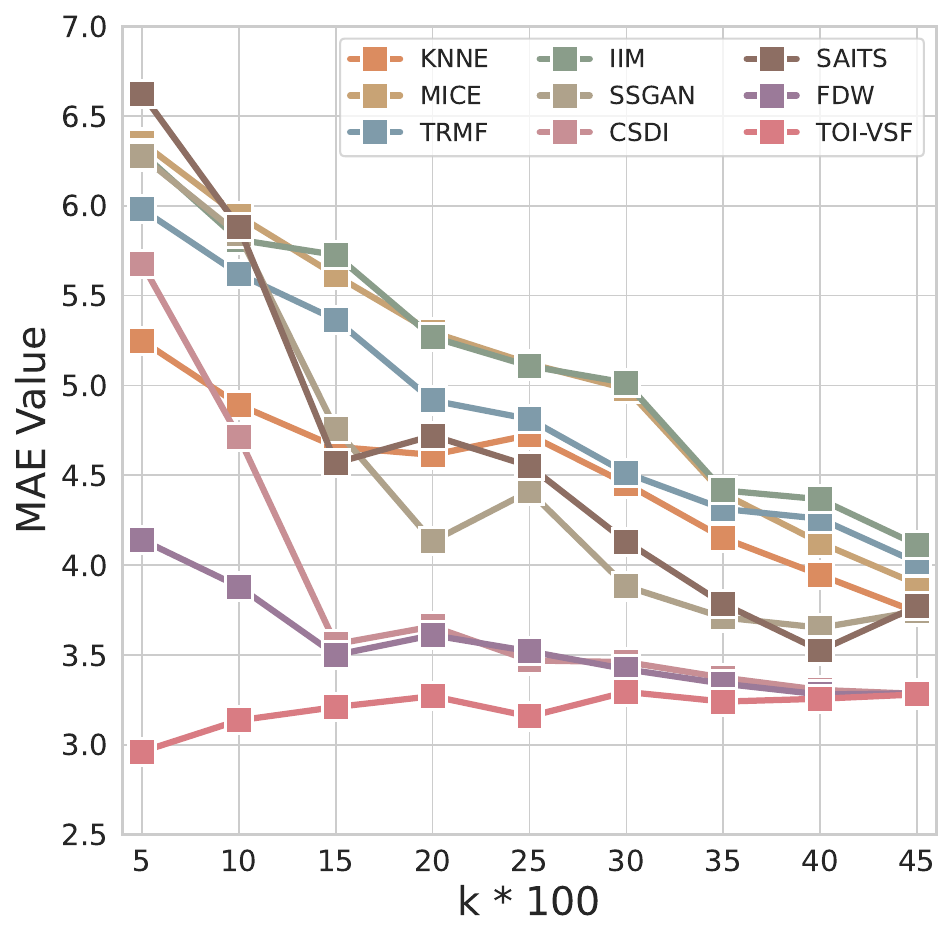}
}
\hspace{-2mm}
\subfloat[RMSE]{
\includegraphics[width=0.49\linewidth]{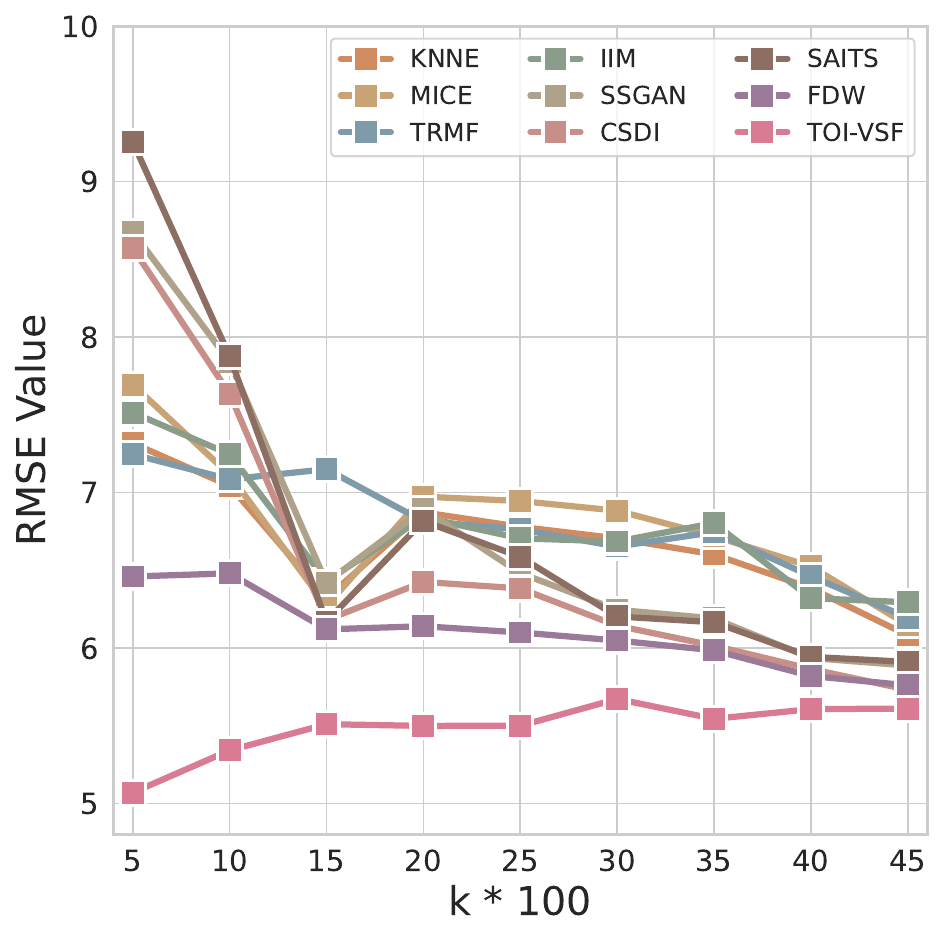}
}
\hspace{-7mm}
\caption{Performance of different $k$ values on VSF, utilizing the backbone MTGNN on the ECG5000 dataset. The lower the results, the better the performance.}
\label{fig: k-ecg}
\end{figure}
\begin{figure}[!th]
\centering
\hspace{-8mm}
\subfloat[MAE]{
\includegraphics[width=0.49\linewidth]{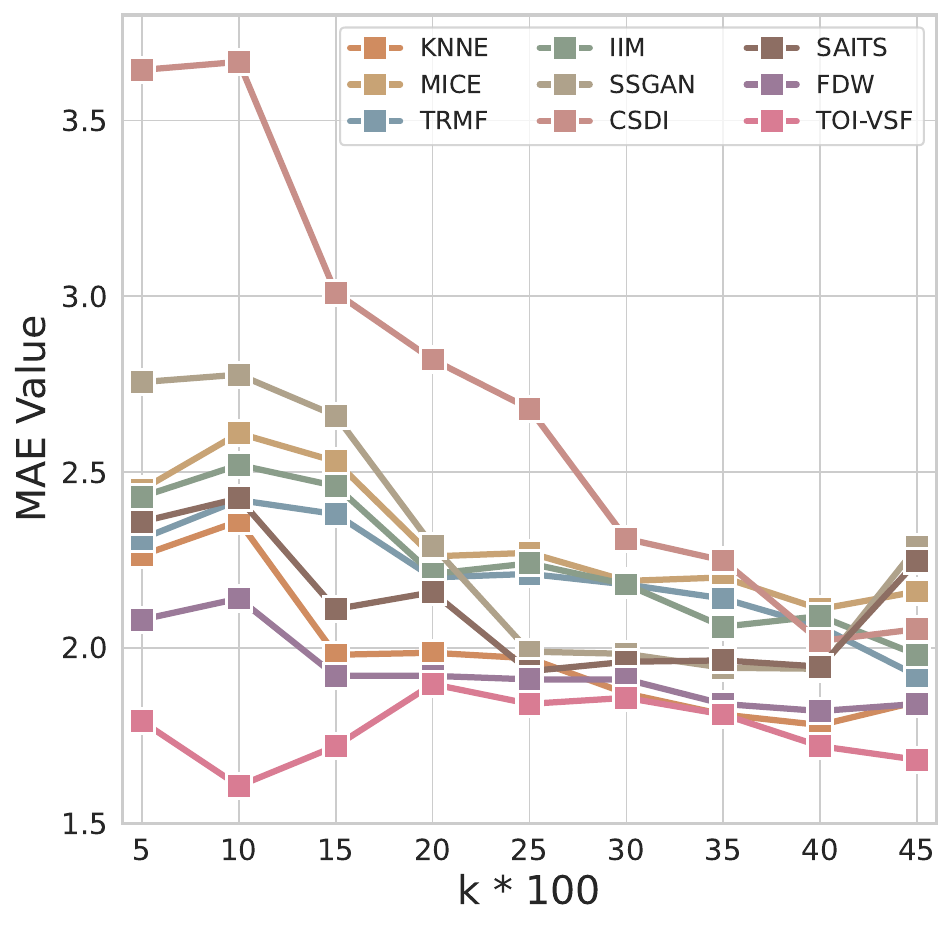}
}
\hspace{-2mm}
\subfloat[RMSE]{
\includegraphics[width=0.49\linewidth]{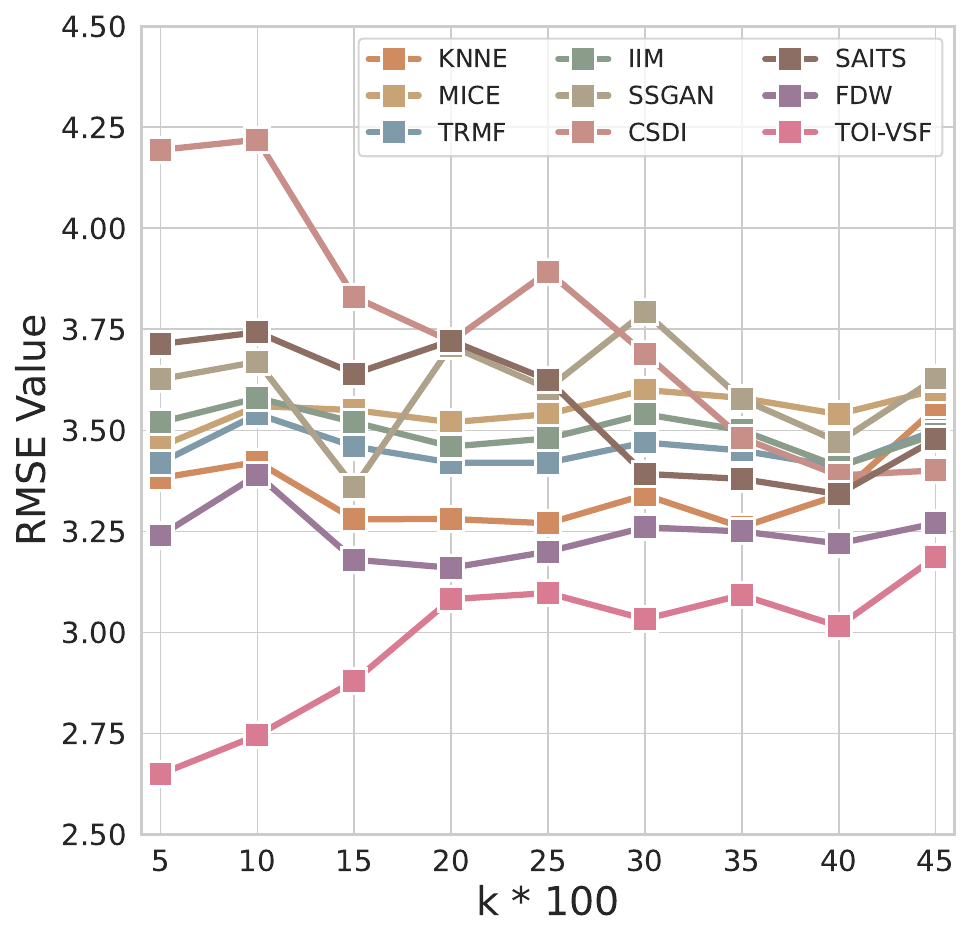}
}
\hspace{-7mm}
\caption{Performance of different $k$ values on VSF, utilizing the backbone MTGNN on the ETTH1 dataset. The lower the results, the better the performance.}
\label{fig: k-etth1}
\end{figure}

\subsection{Impact of Weights Variation of Loss on Forecasting Performance (RQ4)}
In this section, we explore the impacts of the self-supervised variable imputation model and the forecasting model on the VSF task.
Specifically, we set $\alpha + \beta = 1$ and vary the combinations of $\alpha$ and $\beta$
represents the parameters in Equation~\ref{total-loss}, 
and the results are depicted from Figure~\ref{fig:weight-ecg} to Figure~\ref{fig:weight-etth1}.
We can observe that 
when the objective function overwhelmingly favors the forecasting model by only considering the forecasting loss, the model will transform into an end-to-end architecture. 
Consequently, this leads to the generation of data detached from the inherent characteristics of the time series, resulting in a diminished forecasting performance.
In contrast, an exclusive focus on the loss associated with the self-supervised variable imputation model neglects the predictive aspects. This oversight can be detrimental as not all variables positively influence the forecasting outcome—some variables, especially those negatively correlated, can hinder accurate predictions.
Optimal model performance is attained when adopting an intermediate ratio, where parameters $\alpha$ and $\beta$ are approximately equal. 
This balance allows for an effective integration of both imputation and forecasting functionalities, maximizing the strengths of each component.
However, the precise optimal balance may exhibit slight variations depending on the backbones used in the experiments.

\subsection{Comparison with Imputation Methods on VSF (RQ5)}
We also compare our approach to more recent~\cite{domeniconi2004nearest,zhang2019learning,yu2016temporal,tashiro2021csdi,du2023saits,miao2021generative,van2011mice,10.1145/3534678.3539394}
(henceforth referred to as Baseline in Section~\ref{setup}).
Figure \ref{fig: imputation-astgcn} to Figure \ref{fig: imputation-tgcn} demonstrate the forecasting results of five datasets, utilizing four backbones.  
To evaluate the performance of different imputation models, we employ MAE and RMSE as our metrics, with lower values indicating superior performance. 
And among the models tested, TOI-VSF consistently registers the best performance across all datasets.
The results from Figure \ref{fig: imputation-astgcn} to Figure \ref{fig: imputation-tgcn} clearly show the superiority of our self-supervised variable imputation model in accurately completing missing variables so that the forecasting model can generate more informed and reliable outputs.
And the results clearly indicate irrespective of the dataset or backbones chosen, TOI-VSF consistently demonstrates superior performance, thereby substantiating the effectiveness of our proposed task-oriented model. 
To further explain, the absence of variable subset results in the loss of correlation between variables, consequently affecting the performance of data imputation methods.
By effectively completing these missing correlations, TOI-VSF improves imputation accuracy and also enhances the subsequent forecasting performance by preserving the structure and relationships of inputted time series.

\subsection{Performances regarding the Size of Available Variable Subset with Different $k$ (RQ6)}
Here, we explore the impact of the size of the subset of available variables $\Psi_S$ on the final forecasting performance.
Since we set $S$ to be some percent $k$ of all variables $N$, we compare the final forecasting results by setting different values of $k$.
Figure~\ref{fig: k-metrla} to Figure~\ref{fig: k-etth1} show the prediction results of five datasets utilizing the backbone MTGNN.
We observe that as the value of $k$ decreases, the forecasting performance of most baseline methods drops significantly, revealing their limitations in addressing data incompleteness.
In contrast, TOI-VSF consistently maintains stable performance across varying $k$ values, further validating its robustness and reliability. 
Moreover, under different $k$ values, TOI-VSF consistently achieves the best forecasting performance, underscoring the model's superiority in handling complex missing data challenges. 
These findings demonstrate that TOI-VSF not only effectively addresses the challenges posed by missing variables, but also exhibits enhanced robustness and delivers strong performance across various scenarios.



\section{Related Work}
\subsection{Time Series Imputation with Missing Data}
In many cases, the presence of missing data is inevitable. 
Such missing data can significantly impede the performance of forecasting models.
Therefore, model design must consider the impact of missing data on model performance. 
There is a large variety of approaches for time series imputation with missing data~\cite{zainuddin2022timeseries,junger2015imputation,fang2020time,asraf2021missing,hu2024reconstructing}.
Early methods are based on simple value imputation~\cite{asraf2021missing}, like Mean Fill, Median Fill or Plurality Fill.
Later, Neighbor-based method~\cite{song2015turn,sun2020swapping,10.1145/3534678.3539394, domeniconi2004nearest, zhang2019learning} is to search for the neighbors of available data through other attributes for missing data imputation.
For example, KNNE~\cite{domeniconi2004nearest} used $k$-NN algorithm to find the nearest neighbors for missing data case and directly used the corresponding neighbors’ forecast as the test time forecast, weighted via Rank Order Centroid.
FDW~\cite{10.1145/3534678.3539394} involved identifying a subset of variables that closely matches the available variable subset. Nearest neighbors are then used to replace the missing variables, which serve as inputs for the prediction model.
IIM~\cite{zhang2019learning} considered the sparsity and heterogeneity problems present in the data and no longer relies on sharing similar values among $k$ complete neighbors for interpolation, but utilizes the regression results of the individual (not necessarily the same) models learned above.
In addition, machine learning-based methods are proposed, such as, 
MICE~\cite{van2011mice} and TRMF \cite{yu2016temporal}.
In TRMF~\cite{yu2016temporal}, the authors proposed a time-regularized matrix factorization (TRMF) framework using a scalable matrix factorization method to deal with noise and missing value problems in time series data.
MICE~\cite{van2011mice} predicted missing values by using information from existing data and iteratively updated the estimates of missing values.
Recently, deep learning-based methods~\cite{miao2021generative,du2023saits,tashiro2021csdi} mainly deploy Recurrent Neural Network (RNN), for capturing temporal information~\cite{zhang2024spatial}. 
For example, CSDI~\cite{tashiro2021csdi}
proposed Conditional Score-based Diffusion models for Imputation (CSDI), a novel time series imputation method that utilizes score-based diffusion models conditioned on observed data. The model is explicitly trained for imputation and can exploit correlations between observed values.
SAITS~\cite{du2023saits} is trained via a joint optimization approach, learning missing values from a weighted combination of two diagonally masked self-attention (DMSA) blocks. DMSA explicitly captures temporal dependencies and feature correlations between time steps. Meanwhile, the weighted combination design enables SAITS to dynamically assign weights to the representations learned from the two DMSA blocks based on the attention map and missing information.
SS-GAN~\cite{miao2021generative} proposed a novel semi-supervised generative adversarial network model consisting of a generator, discriminator, and classifier. The classifier predicted the labels of the time series data, thereby driving the generator to estimate missing values while being conditioned on the observed components and data labels.

Our approach, different from existing methods, involves the joint training of imputation and forecasting models for task-oriented imputation. 
In contrast, the prevailing literature focuses on imputation for perfect recovery, treating it as a separate process from the forecasting task.


\subsection{Self-supervised Learning} 

The self-supervised learning method acquires a universal feature representation by learning from large-scale unlabeled data. This representation can then be fine-tuned for various downstream tasks, eliminating the need for learning from scratch to meet the specific requirements of different tasks.
Self-supervised learning methods have achieved significant success in natural language processing field~\cite{10.5555/3455716.3455856,elnaggar2021prottrans,baevski2022data2vec} and computer vision~\cite{gidaris2018unsupervised,sermanet2018time,schiappa2023self}. 
Recently, some endeavors have begun to explore the integration of self-supervised learning into time series analysis, which recovers ad hoc positions of the series that are randomly masked~\cite{zhang2023self,zerveas2021transformer,shao2022pre,yoon2019time,poppelbaum2022contrastive,zhang2022self,zhang2024self,ji2023spatio} for capturing inherent spatial and temporal correlations.
For example, ~\cite{zerveas2021transformer} focuses on reconstructing randomly masked positions within the series to capture intrinsic spatial and temporal correlations. 
STEP \cite{shao2022pre} efficiently learns temporal patterns from very long-term historical time series and generates segment-level representations containing contextual information to improve downstream models. 
And \cite{yoon2019time} generates the missing data through a learned embedding space jointly optimized with both supervised and adversarial objectives.

Building on this concept, our work introduces a novel supervised learning framework designed for imputing subsets of variables within the entire time series. 
This task presents a greater challenge compared to the restoration of individual, ad hoc data points in the series.
\section{Conclusion}
In this work, we study the problem of Variable Subset Forecasting (VSF), a critical scenario in time series forecasting where only a small subset of variables are available in the inference phase.
Motivated by the principle of feature engineering that not all variables are beneficial for forecasting, we introduce task-oriented imputation, a new solution by jointly training forecasting and imputation models.  
This method employs forecasting loss to steer the imputation model toward generating variables that enhance the forecasting task.
Specifically, we design the imputation model to generate variables that are most beneficial for the forecasting task, thereby ensuring that the imputed data directly supports the forecasting objective. This task-oriented strategy contrasts with traditional imputation methods, which typically focus on reconstructing missing data without considering the downstream impact on forecasting performance.
In the meantime, we develop a self-supervised imputation model to guarantee the generated variables maintain essential temporal characteristics, thereby maintaining the integrity of the time series while enhancing the forecasting accuracy. 
The extensive results indicate a remarkable improvement over baseline methods, validating the effectiveness of the proposed approach.

\section*{Acknowledgments}
This research is funded by the Science and Technology Development Fund (FDCT), Macau SAR (file no. 0123/2023/RIA2, 001/2024/SKL), the Start-up Research Grant of University of Macau (File no. SRG2021-00017-IOTSC).


\bibliographystyle{IEEEtran}
\bibliography{toivsf}

\end{document}